%% file: main_arxiv.tex
\documentclass[10pt,twocolumn,letterpaper]{article}

\usepackage{cvpr} 
\usepackage[dvipsnames]{xcolor}
\definecolor{cvprblue}{rgb}{0.21,0.49,0.74}
\usepackage[pagebackref,breaklinks,colorlinks,citecolor=cvprblue]{hyperref}
\usepackage{graphicx}
\usepackage{amsmath}
\usepackage[utf8]{inputenc} 
\usepackage[T1]{fontenc}    
\usepackage{url}            
\usepackage{booktabs}       
\usepackage{amsfonts}       
\usepackage{nicefrac}       
\usepackage{microtype}      
\usepackage{multirow}
\usepackage{xspace}
\usepackage{graphicx}
\usepackage{wrapfig}
\usepackage{amssymb}
\usepackage{rotating}
\usepackage{xcolor}[html]
\usepackage{colortbl}

\usepackage[capitalize]{cleveref}
\crefname{section}{Sec.}{Secs.}
\Crefname{section}{Section}{Sections}
\Crefname{table}{Table}{Tables}
\crefname{table}{Tab.}{Tabs.}

\newcommand{\name}{SMPLer-X\xspace}
\newcommand{\Fig}{Fig.\xspace}
\newcommand{\Tab}{Table\xspace}
\newcommand{\Sec}{Sec.\xspace}

\def\confName{CVPR}
\def\confYear{2024}
\title{AiOS: All-in-One-Stage Expressive Human Pose and Shape Estimation}
\author{%
  Qingping Sun\thanks{Equal contributions.} $^{,1,2}$, Yanjun Wang$^{*,1}$, Ailing Zeng$^{3}$, Wanqi Yin$^{1}$, Chen Wei$^{1}$, \\Wenjia Wang$^{5}$, Haiyi Mei $^{1}$, Chi-Sing Leung$^{2}$, Lei Yang$^{1,5}$, Ziwei Liu$^{4}$, Zhongang Cai\thanks{Corresponding author.} $^{,1,4,5}$ \\
  $^{1}$ SenseTime Research \hspace{2mm}
  $^{2}$ City University of Hong Kong  \\
  $^{3}$ International Digital Economy Academy (IDEA) \\
  $^{4}$ S-Lab, Nanyang Technological University \hspace{2mm} 
  $^{5}$ Shanghai AI Laboratory  \\
  {\small $^{*}$ Equal Contributions, $^{\dagger}$ Corresponding Author} \\
  {\url{https://ttxskk.github.io/AiOS/}}
}

\begin{document}

\twocolumn[{%
	\renewcommand\twocolumn[1][]{#1}%
	\maketitle
	\begin{center}
		\newcommand{\teaserwidth}{1\linewidth}
		\centerline{
			\includegraphics[width=1\linewidth,clip]{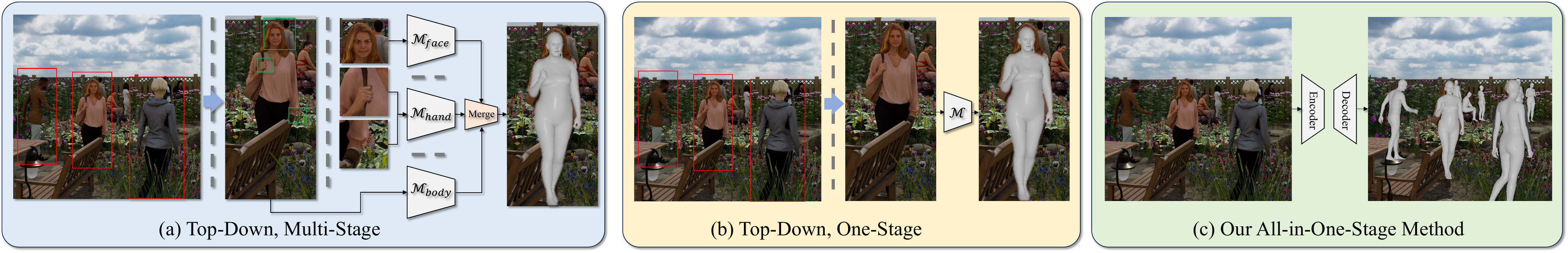}
		}
		\vspace{-1ex}
		\captionof{figure}{\label{fig:teaser}A comparison of existing methods in EHPS. (a) Top-down, multi-stage methods, typically use detectors to detect humans, then use different networks to regress body parts on cropped images. (b) Top-down, one-stage methods, use only one network for regression but still require detectors and rely on the cropped image. (c) Our all-in-one-stage pipeline, end-to-end human detection, and regression on full frame.}
	\end{center}%
}]
  

\input{sections/0_abstract}
\input{sections/1_introduction}

\input{sections/2_related_works}

\input{sections/3_method}

\input{sections/4_experiments}

\input{sections/5_conclusions}

\clearpage
{\small
\bibliographystyle{ieee_fullname}
\bibliography{main}
}

\clearpage
\maketitlesupplementary
\appendix

\input{sections/A_overview}

\input{sections/B_datasets}

\input{sections/C_experiment_1}

\input{sections/F_experiments_3}


\clearpage
\end{document}

%% file: sections/0_abstract.tex
\begin{abstract}

Expressive human pose and shape estimation (a.k.a. 3D whole-body mesh recovery) involves the human body, hand, and expression estimation. Most existing methods have tackled this task in a two-stage manner, first detecting the human body part with an off-the-shelf detection model and inferring the different human body parts individually. Despite the impressive results achieved, these methods suffer from 1) loss of valuable contextual information via cropping, 2) introducing distractions, and 3) lacking inter-association among different persons and body parts, inevitably causing performance degradation, especially for crowded scenes. To address these issues, we introduce a novel all-in-one-stage framework, AiOS, for multiple expressive human pose and shape recovery without an additional human detection step. Specifically, our method is built upon DETR, which treats multi-person whole-body mesh recovery task as a progressive set prediction problem with various sequential detection. We devise the decoder tokens and extend them to our task. 
Specifically, we first employ a human token to probe a human location in the image and encode global features for each instance, which provides a coarse location for the later transformer block. Then, we introduce a joint-related token to probe the human joint in the image and encoder a fine-grained local feature, which collaborates with the global feature to regress the whole-body mesh.  
This straightforward but effective model outperforms previous state-of-the-art methods by a 9\% reduction in NMVE on AGORA, a 30\% reduction in PVE on EHF, a 10\% reduction in PVE on ARCTIC, and a 3\% reduction in PVE on EgoBody. 
\end{abstract}

%% file: sections/1_introduction.tex
\section{Introduction}

Expressive human pose and shape estimation (EHPS) \footnote{EHPS is used interchangeably with 3D whole-body human mesh recovery in this work} is a rapidly developing area. It plays an important role in human understanding and has broad applications in the animation, gaming, and streaming industries.
Unlike human pose and shape estimation (HPS), which focuses solely on the human body, EHPS is designed to jointly estimate human body poses, hand gestures, and facial expressions from the image. 

In mainstream studies, the common approaches involve utilizing parametric human models, such as SMPL-X~\cite{smplx}, to represent the articulated mesh model of a human and to regress the parameters for each body part. Drawing from research experience in single-part estimation, such as body pose and shape estimation~\cite{hmr, spin, pare, pymaf, li2020hybrik, cliff, zolly, tore, hybridhmr, wang2023learning}, existing methods~\cite{expose, frankmocap, pixie, hand4whole, hybrikx, smplerx, robosmplx} employ a multi-stage paradigm. As shown in \Fig\ref{fig:teaser}a), the process begins by cropping the body parts using bounding boxes detected either by off-the-shelf detection models or provided via ground truth annotations. Following this, distinct models are utilized for the separate reconstruction of each individual body part.

Obviously, this design compromises both complexity and accuracy. The images are processed multiple times with each model. The separate parts model blocks the inter-part, inter-human connection and brings inconsistent poses and unnatural artifacts at the connected joints. Recently, OSX~\cite {osx} and SMPLer-X~\cite{smplerx} discard part experts and regress the model in a holistic manner, which alleviates the artifacts. Their paradigm can be abstract to \Fig\ref{fig:teaser}b), however, they still need to be given a bounding box to crop the image. While their benchmarks show promising results, the accurate ground truth bounding boxes are not attainable in real-world scenarios. RoboSMPLX~\cite{robosmplx} has demonstrated that the performance drops significantly under noisy boxes. Moreover, CLIFF~\cite{cliff} points out that the cropping operation discards the location information, which degrades the performance.

A direct solution to address the challenges posed by the multi-stage paradigm is to utilize a one-stage framework that directly recovers EHPS from the entire image without requiring additional boxes for cropping. However, current one-stage methods~\cite{romp,bev, trace} are proposed for HPS. Both of them use a body center heatmap and mesh parameter map to represent the potential human location and corresponding features. Relying solely on these human-centered global features is insufficient for achieving accurate part-wise regression. Although numerous two-stage HPS methods~\cite{pare, decomr} that extract local features in various ways, it is non-trivial to extend to a one-stage model, as most of the representations, like part-attention maps, are designed for a single person.

In order to tackle the above challenges, we have proposed the first All-in-One-Stage (AiOS) EHPS method. This novel approach is capable of predicting every individual present in an image solely based on a single image input without any additional requirements.  Inspired by the achievement of DETR-based~\cite{detr} methods in various vision tasks~\cite{edpose, motr, hoitransformer, cdn, yang2023neural}, we designed our pipeline in a DETR~\cite{detr} style with image feature encoder and various location-aware decoders. We tailored different queries and association strategies to progressively guide the decoder to perceive global and local human features from the entire image. 

Three key design features distinguish the AiOS model. \textbf{First}, it is built upon DETR structure, with a CNN backbone, transformer encoders, and decoders, and progressively detects human and decodes person features in an end-to-end manner. \textbf{Second}, we introduced the "Human-as-Tokens" design, where humans are conceptualized as a collection of box tokens and joint tokens. With different supervision and location cues, these tokens aggregate both global and local feature representations with cross-attention for enhanced model accuracy in diverse scenarios. \textbf{Third}, using self-attention and cross-attention mechanisms in our model allows for an in-depth analysis of inter-human and intra-human relationships, enhancing performance in crowded and occlusion-heavy environments.

Extensive experiments show that our proposed model has overpass state-of-the-art (SOTA) methods that utilize ground truth bounding boxes and also SOTA methods when the bounding box is not given. Further, our bounding box is accurate enough to improve the other two-stage methods on the AGORA benchmark. 

In summary, our contributions are i) The first one-stage method for EHPS that eliminates the need for extra detection networks; ii) A unified framework to integrate local and global features for whole-body regression; iii) SOTA performance on mainstream benchmarks without ground truth bounding boxes.

%% file: sections/2_related_works.tex
\section{Related Work}

\subsection{Expressive Human Mesh Recovery Methods}
EHPS focuses on reconstructing the mesh of the human body, hands, and face from monocular images. Pioneering research in this domain introduced whole-body parametric models such as SMPL-X~\cite{smplx}.
With advancements in regression techniques for the human body, hands, and face, early studies adopted multi-stage solutions~\cite{PavlakosGeorgios2020expose, frankmocap, pixie, hand4whole}. They independently recover body pose, hand pose, and facial expressions from cropped images before integration. However, these multi-stage methods often produce artifacts at joint intersections and present complex network designs.

Given the recent surge in whole-body datasets~\cite{yang2023synbody,cai2022humman, bedlam, GTAHuman, agora}, many approaches have transitioned to a holistic paradigm. OSX~\cite{osx} presents a groundbreaking one-stage method, eliminating part-specific experts and cropped image regression. SMPLer-X~\cite{smplerx} further amplifies one-stage methods utilizing large vision models and extensive datasets. However, they still rely on bounding boxes for image cropping. Despite their precision with ground truth bounding boxes, performance degrades under detected boxes~\cite{robosmplx}. 

\subsection{One-Stage Human Mesh Recovery Methods}
Most of the existing HPS methods~\cite{hmr, zolly, decomr, smoothnet, deciwatch, graphcmr, meshgraphormer, fastmetro, tore} are multi-staged. Although these methods preserve relatively high-resolution images and generally have higher accuracy, they neglect other information in the full frame, including inter-person occlusions and individual positions~\cite{cliff}. To address these limitations, ROMP~\cite{romp} first proposed to recover humans from an entire frame. It locates the human locations from a body center heatmap and indexes the corresponding features from the feature map to regress all human meshes. Furthermore, BEV~\cite{bev} extends the 2D heatmap to 3D by incorporating a bird-eye-view. It enables the model to discern 3D relative positions within the frame. TRACE~\cite{trace} further achieved simultaneously tracking humans and predicting camera motions with added motion maps. 
However, these center-map-based methods often distill the human into a single vector on the feature map and recover the human pose and shape based on this global feature. We reckon that this representation is insufficient for the EHPS task, particularly given that hand pose and expression require more fine-grind local features for accurate regression.

%% file: sections/3_method.tex
\section{Method}

\begin{figure*}[t]
  \centering
   \includegraphics[width=1\linewidth]{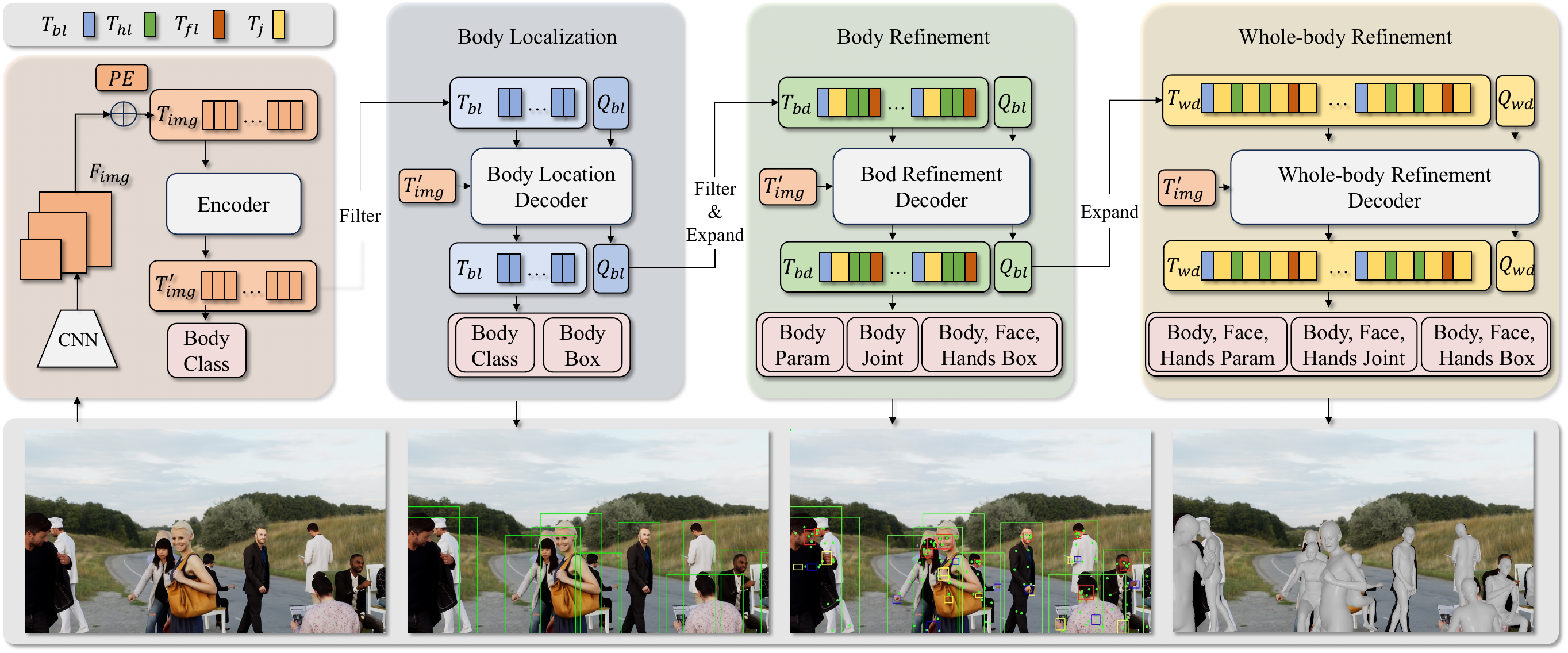}
  \vspace{-0.25in}
  \caption{\textbf{Pipeline overview}. AiOS performs human localization and SMPL-X estimation in a progressive manner. It is composed of (1) the body localization stage that predicts coarse human location; (2) the Body refinement stage that refines body features and produces face and hand locations; (3) the Whole-body Refinement stage that refines whole-body features and regress SMPL-X parameters.}
  \label{fig:pipeline}
  \vspace{-4mm}
\end{figure*}

\subsection{Motivation}
For EHPS, using cropped images presents significant problems. The cropping discards the location information~\cite{cliff}, and inaccurate bounding boxes may lead to missing body parts, negatively impacting performance. In crowded scenes, cropping struggles to distinguish individual humans, with parts from others intruding into the frame, leading to errors in human part detection and regression. Especially when people overlap significantly, the model struggles to differentiate them due to unclear bounding boxes. Furthermore, the detectors used are typically trained on general object detection datasets and are not specifically designed for human detection, adding to these difficulties.

To tackle these problems, we introduced the AiOS, the first fully end-to-end network for EHPS. Abandoning the uncertain assumption of box-as-subject, our model leverages feature tokens and position queries for more precise human localization. We've developed a cohesive approach that combines global and local feature representations for accurate regression. To handle crowded scenarios and enhance the separation of human figures, our model employs attention mechanisms to establish intricate relationships between different body parts and between multiple individuals.

\subsection{Preliminaries}
\noindent\textbf{SMPL-X.} We use 3D parametric model SMPL-X ~\cite{smplx} to study EHPS. It utilizes a set of parameters to model body, face, and hands geometries. Specifically, our model estimates pose parameters $\theta \in \mathbb{R}^{53\times3}$, which include body poses $\theta_{body} \in \mathbb{R}^{22\times3}$, left hand poses $\theta_{lhand}\in \mathbb{R}^{15\times3}$, right hand poses $\theta_{rhand}\in \mathbb{R}^{15\times3}$, and jaw poses $\theta_{jaw} \in \mathbb{R}^{1\times3}$. Additionally, it estimates shape parameters $\beta \in \mathbb{R}^{10}$, and facial expression parameters $\psi \in \mathbb{R}^{10}$. We use the joint regressor ${J}$ to obtain the 3D joint from the parameters by $J(\mathcal{M}(\beta,\theta, \psi))$, where $\mathcal{M}$ is the SMPL-X function.

\subsection{Overview}
AiOS includes the backbone and transformer encoder-decoder structures. It has three steps, 1) localize coarse body location and extract global features of the body; 2) refine body location, extract body local features, localize coarse hands and face locations and extract global features of the hands and face; 3) refine hand and face location, extract local features for whole body.

\noindent\textbf{Backbone.} AiOS utilizes the ResNet-50~\cite{he2016deep} to extract a multi-scale feature maps $F_{img}$, which provide features from detailed to holistic. 

\noindent\textbf{Encoder.} As our task needs more than local associations, we utilize a standard transformer encoder~\cite{vaswani2017attention} for long-distance relations. To transform the CNN-based feature map into a transformer-compatible feature vector, we flatten the multi-layer feature maps along their spatial dimensions and concatenate them.  The flattened feature is added with position encodings $PE\in \mathbb{R}^{M \times D}$ to derive the image feature token $T_{img} \in \mathbb{R}^{M \times D}$, where $M$ represents the total length of the image feature token. We fed $T_{img}$ a transformer encoder, which produces the refined image feature tokens $T^{'}_{img}$, serving as a reference for cross-attention in the decoder. Utilizing $T^{'}_{img}$, a feed-forward network (FFN) is applied to classify each token as a human token. Following the approach in DINO~\cite{zhang2022dino} and ED-Pose~\cite{edpose}, we filter based on the classification score and retain the top $M_h = 900$ tokens. These tokens serve as candidate human body localization tokens $T_{body} \in \mathbb{R}^{M_h \times D}$, and they also function as the input for the subsequent decoders.

\noindent\textbf{Generic Decoder.} Similar to PETR~\cite{petr} and ED-Pose~\cite{edpose}, which extend the deformable decoder~\cite{deformable} to 2D human body-only pose estimation, AiOS extends the deformable decoder to 3D whole-body mesh recovery. It mainly has three inputs, image content tokens $T^{'}_{img} \in \mathbb{R}^{M \times D}$, object content tokens $T\in \mathbb{R}^{M_h \times D}$ and object position queries $Q\in \mathbb{R}^{M_h \times 4}$.
Utilizing this decoder, our model can automatically probe the suitable global and local features around the body parts for each human conditioned by various queries. We will introduce our key decoder designs in the following sections.

\subsection{Naive AiOS}
Drawing inspiration from ROMP~\cite{romp}, we extend the DETR structure~\cite{detr} to EHPS and progressively regress SMPL-X parameters. Specifically, we follow DAB-DETR~\cite{liu2022dab} and introduce the location queries to probe the body, face, and hands-related features, guided by bounding boxes $(x, y, w, h)$, that considers both the location and size of each body part boxes. The model first extracts features related to the body using body box location queries and refines them through the body-location decoder. Subsequently, they are expanded to include hands and face queries and leverage the whole-body-location decoder to extract whole-body features. 

\noindent\textbf{Body-location Decoder.} The first two decoders are body-centric, and the input object content tokens $T$ are the body location tokens $T_{bl}$. We derive body location query $Q_{bl}$ with FFN from the corresponding $T_{bl}$. The decoder first associates the body location tokens and updates them by the self-attention mechanism. Then, the decoder takes image tokens $T^{'}_{img}$ as the value and the updated body location tokens as the query for cross-attention, and the $Q_{bl}$ acts as an indicator, which is used to aggregate the information focusing on the corresponding body area. After that, the body location tokens $T_{bl}$ and body location queries $Q_{bl}$ are refined with the decoder.

We estimate the body bounding box with an FFN from $T_{bl}$, which is supervised by $L_{box}$. This supervision makes sure the tokens aggregate global information of the human. Similar to the encoder, we classify the output $T_{bl}$ with an FFN on whether it is a token representing a human. The classification results from $T_{bl}$ are supervised with classification loss $L_{cls}$. At the end of the second decoder, we downsample the body tokens again to $M_b = 100$ to further distill potential human tokens and lower the computational complexity.

\noindent\textbf{Whole-body-location Decoder.} The latter four decoders of naive AiOS jointly consider whole-body information and their association. With the body location tokens from the previous step, we expand them to hands and face location tokens with learnable embedding. We first broadcast the given embedding $E_{bl} \in \mathbb{R}^{D}$ and add it to the body location token $T_{bl}\in \mathbb{R}^{M_b \times D}$. After that, we obtained hand location tokens $T_{lhl}$, $T_{rhl}$, and face location tokens $T_{fl}$, which have the same shape as $T_{bl}$. Then we concat them into a whole-body token $T_{full} = [T_{bl}, T_{lhl}, T_{rhl}, T_{fl}]$. Similarly, the whole-body location queries $Q_{full}$ are expanded from $Q_{bl}$ with learnable embeddings.

The decoders use a self-attention module to explore inter-part and inter-human relations and then extract each part's features around their bounding boxes with a conditioned cross-attention module. 
We utilize an attention mask to ensure that the bounding boxes for each person's hands and face are associated only with their own and others' body bounding boxes. As our model is already capable of recognizing each person's body in the first two stages, this specific attention mechanism allows for more accurate identification of body parts in crowded scenes.
We provide an illustration of the attention mechanism in the Supplementary Material.

We regress body bounding boxes from $T_{bl}$, face boxes from $T_{fl}$ and hand boxes from $T_{rhl}$, $T_{lhl}$, and supervise them with $L_{box}$. We regress different part's parameters from the refined whole body $T_{full}$ tokens.  The parameters are supervised with SMPL-X loss$L_{smplx}$, which includes parameter loss $L_{param}$, 3d keypoints loss $L_{kp3d}$, and the 2d keypoints reprojection loss $L_{kp2d}$. 

\subsection{AiOS}
Previous methods~\cite{romp, bev} have shown that regressing multi-person body meshes from global features alone can achieve impressive results, but in EHPS, relying on global information alone is insufficient. The model should also consider local information to obtain a detailed context of the whole-body regression. Therefore, to elevate the model's ability, we introduce joint-related tokens and their corresponding queries to our model. Combined with location tokens, the AiOS expresses human context in multilevel. We will further regress the SMPL-X parameter on this well-rounded feature group. 
Specifically, we adopt a progressive detection and decoding strategy. The first two layers are body-location decoders same as our naive design, which outputs coarse human location. Further, two layers of body-refinement decoders utilize body joint tokens to enrich local body features and estimate rough hand and face location simultaneously on the basis of human location. At last, two layers of whole-body-refinement decoders extract whole-body local features with extra hands and face joint tokens. 

\noindent\textbf{Body-refinement Decoder.} This decoder is built on body-location decoders in naive AiOS. In detail, we expand body joints tokens, hands location tokens, and face location tokens. We adopt the learnable-embedding $E_{bj}\in \mathbb{R}^{17 \times D}$ to expand body joint tokens $T_{bj}\in \mathbb{R}^{M_b \times 17 \times D}$ from box location tokens, and then we obtain detailed body token set $T_{bd} = [T_{bl}, T_{bj}, T_{lhl}, T_{rhl}, T_{fl}]$. Note that we use an attention mask to limit the joint attention within its subject as inter-joint attention among different subjects brings no incremental but much higher computation complexity. 


The $T_{bd}$ are refined with layers of decoders. Within each layer, similar to naive AiOS, we regress bounding boxes of body parts from their location tokens and supervise them with $L_{box}$. Further, we regress body joint location from $T_{bj}$ and supervise them with $L_{j2d}$, helping these joint tokens learn the local human features. Different from Naive AiOS, in this stage, we regress SMPL-X body parameters based on $T_{bl}$, $T_{bj}$.  We use $L_{smplx}$ to supervise the body parameter, helping to refine the body-related tokens representing more accurate body features.

\noindent\textbf{Whole-body-refinement Decoder.} This decoder further expands the face and hand joint tokens. Similarly, we use embedding $E_{lhj}$, $E_{rhj}$, and $E_{fj}$ to expand $T_{lhl}$, $T_{rhl}$, and $T_{fl}$ to $T_{lhj}$, $T_{rhj}$, and $T_{fj}$, respectively. At this stage, the model forms the complete tokens that represent a human $T_{wd} = [T_{bl}, T_{bj}, T_{lhl}, T_{lhj}, T_{rhl}, T_{rhj}, T_{fl}, T_{fj}]$. We regress whole-body joint location from $T_{bj}, T_{lhj}, T_{rhj}, T_{fj}$ and supervise them with $L_{j2d}$. 

Based on $T_{wd}$, we utilize FFN to regress box location from $T_{bl},T_{lhl}, T_{rhl}, T_{fl}$ and supervised with $L_{box}$. We also regress whole-body joint location from $T_{bj}$, $T_{lhj}$, $T_{rhj}$, and $T_{fj}$, and supervise them with $L_{j2d}$. Finally, we estimate SMPL-X body, hands, and face parameters from body, hand, and face-related tokens, respectively, and supervise whole-body parameters with $L_{smplx}$. 

\noindent\textbf{Overall Loss Functions.} The overall loss function is the sum of all the losses at each stage. Please refer to the Supplementary Material for the details.

%% file: sections/4_experiments.tex
\section{Experiment}

\subsection{Experimental Setup}
Due to the page limit, we put the detailed experiment setup, implementation, and partial quantitative and qualitative comparison with SOTA methods in the Supplementary Material.

\noindent\textbf{Datasets.} 
AiOS is trained on the multi-person datasets AGORA~\cite{agora}, BEDLAM~\cite{bedlam}, and COCO~\cite{coco}, and single-person datasets UBody~\cite{osx}, ARCTIC~\cite{arctic}, and EgoBody~\cite{egobody}. We evaluate it on AGORA, UBody, EHF~\cite{smplx}, ARCTIC~\cite{arctic}, Egobody~\cite{egobody}, and BEDLAM~\cite{bedlam}.

\input{tables/agora_test}
\input{tables/agora_smpl_test}

\noindent\textbf{Implementation.}
The training is conducted on 16 V100 GPUs, with a total batch size of 32. We first train our model for 60 epochs on AGORA, BEDLAM, and COCO. We finetune it for 50 epochs on all train datasets.

\noindent
\textbf{Evaluation metrics.}
Following the previous EHPS methods~\cite{hand4whole, osx, smplerx}, we report Procrustes Aligned per-vertex position error (PA-MPVPE) and the mean per-vertex position error (MPVPE) across all benchmarks. In AGORA Leaderboard, we report mean vertex error (MVE), mean per-joint position error (MPJPE) for pure reconstruction accuracy; F Score, precision, recall for detection accuracy; Normalized mean vertex error (NMVE) and normalized mean joint error (NMJE) that considered regression accuracy with detection accuracy. All metrics are reported in millimeters (mm).

\subsection{Quantitative comparison with SOTA}

In \Tab~\ref{tab:agora_test}, we compare AiOS with the SOTA methods on the AGORA test set. The results are provided by the leaderboard~\footnote{{https://agora-evaluation.is.tuebingen.mpg.de/}} with their bounding boxes on the upper part of the table. We also feed our estimated bounding boxes to OSX~\cite{osx} and SMPLer-X~\cite{smplerx} on the lower part, which helps to verify our model's localization quality. 

For a fair comparison with the SOTA methods, we utilize a threshold of 0.5 to filter the detected samples with lower confidence, which generally have severe occlusions. As shown on the upper part of \Tab~\ref{tab:agora_test}, our model's NMVE and NMJE greatly surpass the current SOTA method SMPLer-X. This observation proves that our one-stage pipeline achieves the best overall quality, combining localization and reconstruction. In terms of pure reconstruction quality, our model also achieves SOTA performance with a relatively accurate detection result on MVE and MPJPE. While BEDLAM~\cite{bedlam} excels in face and hand reconstruction, its recall performance is comparatively low, omitting some instances for evaluation.

On the lower-part comparison, we lower the detection threshold to 0.3,  which has higher recall than any current results, allowing more hard cases to be detected. We feed the same bounding boxes to the OSX and SMPLer-X, and their performance on whole-body MVE improves compared with the results reported in the original paper (122.8 to 121.3 for OSX, 99.7 to 98.3 for SMPLer-X) even with a higher recall. This indicates that improvement is achieved not by filtering out hard cases but by providing high-quality bounding boxes. This finding proves that the current two-stage method is sensitive to bounding box quality, and using the ground truth box to crop images in other benchmarks is biased from real use cases.
Notably, under this bounding box setting, AiOS is still much higher than the current SOTA OSX and comparable with the foundation model SMPLer-X L20.

As the first one-stage method in EHPS, we cannot find relevant one-stage methods for a fair comparison. Therefore, similar to H4W~\cite{hand4whole}, we compare the results of our body part with existing body-only methods, which can be broadly categorized into top-down methods~\cite{hmr,pymaf,pare,cliff, li2020hybrik,propose,pliks, niki} and one-stage methods~\cite{romp,bev}, on the AGORA SMPL test set.
Specifically, we downsample the SMPL-X mesh estimated by AiOS to the SMPL~\cite{smpl} mesh using official tools~\cite{smplx} and then measure MVE and NMVE. We use the J-regressor to regress joints from the downsampled SMPL mesh to measure NMJE and MPJPE.

As shown in \Tab~\ref{tab:agora_smpl}, even though AiOS is designed for EHPS, it still outperforms ROMP~\cite{romp} and BEV~\cite{bev}, with a notable improvement in NMVE of 43\% (from 108.3 mm to 61.2 mm) and an NMJE enhancement of 40\% (from 113.2 mm to 68.0 mm). It is worth noting that we do not deliberately fine-tune our model exclusively on AGORA.

\noindent
\textbf{Single datasets.} We compare UBody in \Tab~\ref{tab:ubody}, EHF in \Tab~\ref{tab:ehf}. Note that the other methods utilize ground-truth bounding boxes. Without any given bounding boxes, our model achieves SOTA performance on real-life datasets. 
\input{tables/ubody}
\input{tables/ehf}

\begin{figure*}[h]
    \centering
  \includegraphics[width=1.0\linewidth]{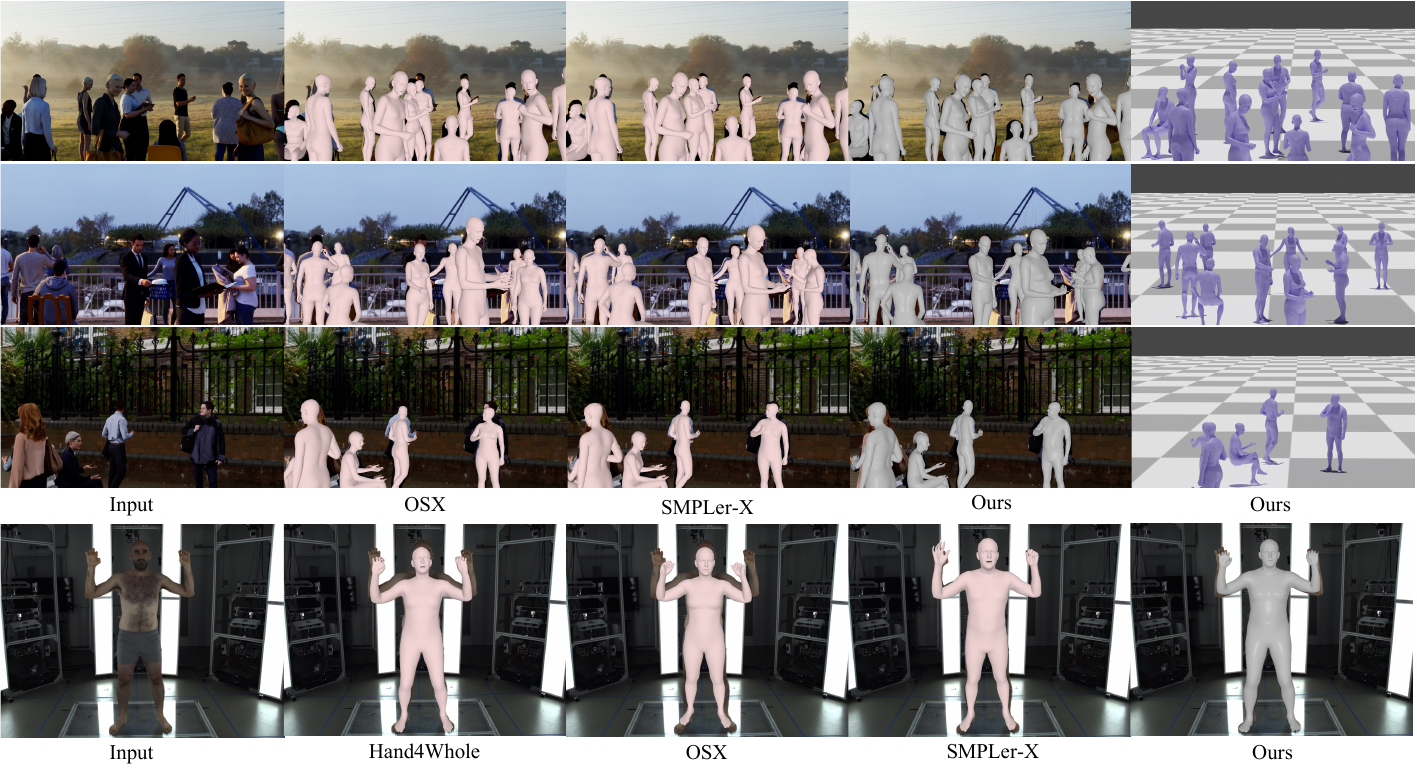}
  \vspace{-20pt}
    \caption{Comparison of current SOTA methods ~\cite{smplerx,hand4whole,osx} with our AiOS model. The upper part is visualization results on AGORA~\cite{agora}, and the lower is EHF test~\cite{expose}.}
    \label{fig:datasetsota}
    \vspace{-2mm}
\end{figure*}

\begin{figure*}[h]
    \centering
  \includegraphics[width=1\linewidth]{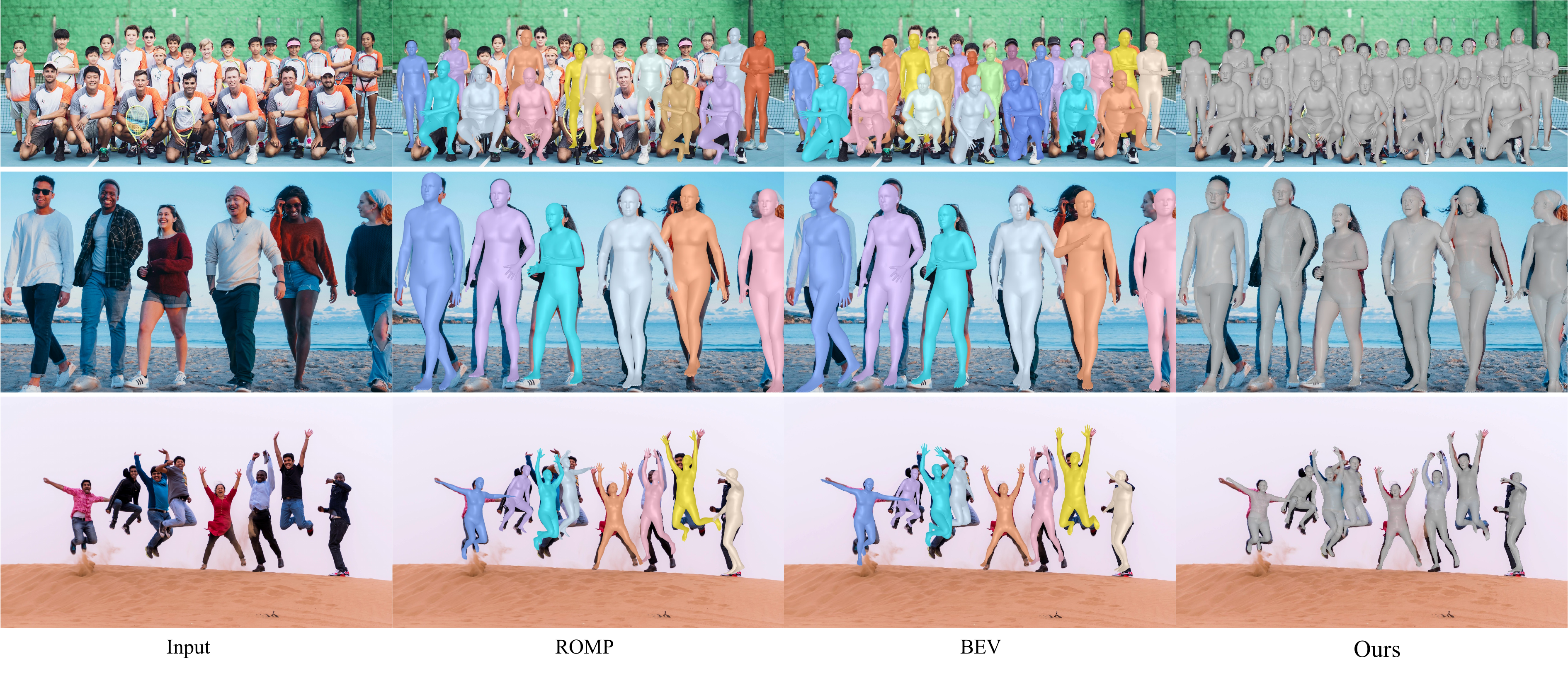}
  \vspace{-20pt}
    \caption{{Visual comparisons with SOTA one-stage HPS methods~\cite{romp, bev} on the Internet data\protect\footnotemark.} 
    }
    \label{fig:romp}
    \vspace{-4mm}

\end{figure*}

\subsection{Qualitative comparison with SOTA}

We perform a qualitative comparison with current SOTA methods on AGORA and EHF. To overlay the results onto the image, we apply an affine transformation for the two-stage methods that use images cropped by ground truth boxes. In contrast, our method can be directly overlaid on the image. Further, with accurate betas estimation, we are able to recover the depth order, as shown in the \Fig\ref{fig:datasetsota}. We achieve comparable visual quality in both scenes, proving our model's accuracy.

We further perform a qualitative comparison with SOTA one-stage methods~\cite{romp, bev}. As shown in \Fig\ref{fig:romp}, while ROMP and BEV can achieve decent results for body reconstruction in multi-person scenarios, they are limited by the constraints of the SMPL~\cite{smpl} model, preventing them from reconstructing detailed hand gestures and facial expressions.

\begin{figure}[h]
    \centering
  \includegraphics[width=0.9\linewidth]{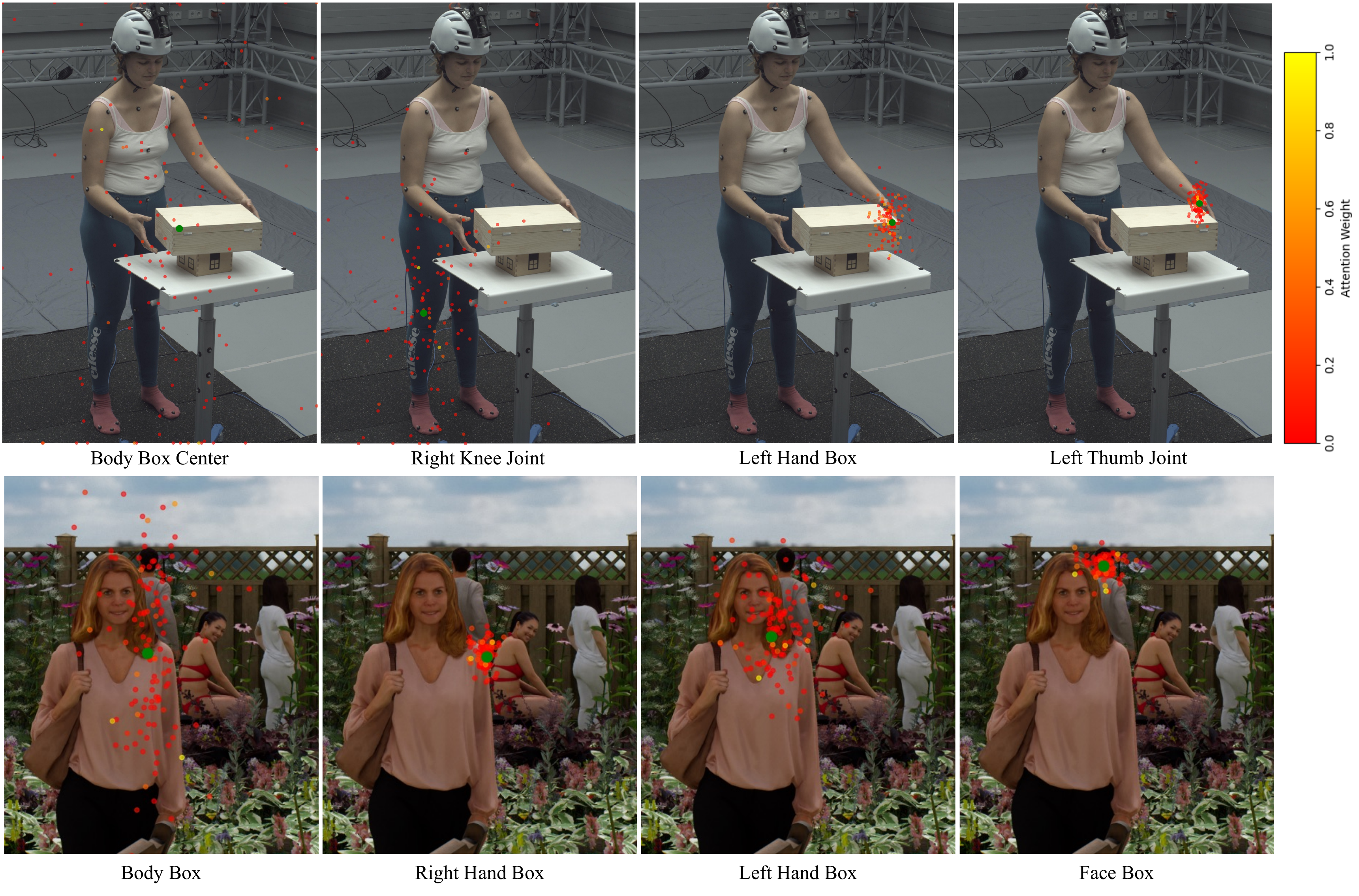}
    \caption{\textbf{Attention Visualization.} The green dots represent the location of the reference point, and the red dots are the sampling points. }
    \label{fig:attention}
    \vspace{-10pt}
\end{figure}

\subsection{Ablation Study}
In this subsection, we analyze the effectiveness of the proposed components in detail. All experiments are conducted on the AGORA validation set. 

\footnotetext{https://www.pexels.com/}

\noindent\textbf{Analysis of the naive AiOS and full AiOS.}
Whole-body mesh recovery requires attention to both small-scale gestures, expression details, and large-scale pose details. To validate the effectiveness of our joint-guided local feature query, we compared naive AiOS and full AiOS models across the benchmarks. The table shows that even the naive setting achieves comparable performance with SOTA methods, indicating our one-stage pipeline, which treats EHPS as a progressive set prediction problem with various sequential detections following the DETR, is ideal for SMPL-X parameter regression.  On this solid base, the full AiOS consistently achieves higher accuracy on all parts of the human, and the increment on the whole-body aspect is especially outstanding. Since the body tends to have a relatively higher area on the image, adding joint queries to the body provides a large number of local features for reference, while for smaller areas like face and gestures, the difference between global and local features is not that obvious. However, adding the local joints feature overall brings more comprehensive features.

\noindent
\textbf{The Scheme of the SMPL-X supervision.}
In this part, we investigate how to supervise different tokens. For our original AiOS, we don't supervise the SMPL-X parameter in the first stage, as we want the model to focus on body localization. In the second stage, we don't supervise hands and face for the same reason, but supervise SMPL-X body parameters as we have detailed body feature tokens. And we supervise the whole body parameter at the third stage. 
In the first ablation setting, we add body parameter supervision in every stage and hand and face supervision in the second stage, meaning every stage has SMPL-X supervision. In the second setting, we remove the SMPL-X body supervision in the second stage so that the model will be only supervised by SMPL-X in the last stage. As shown in \Tab~\ref{tab:abl2}, a comparison between AiOS and all stage settings shows adding SMPL-X parameters when the location is not properly refined will hinder the model's performance. Comparing the AiOS and 3rd stage setting shows the design of gradually whole-body estimation from body to whole-body increased performance.
\input{tables/ablation_2}
\noindent\textbf{The association between the human body, hands, and face.}
We focus on the self-attention relations on this part. Our stock design allows free attention among body, face, and hand location tokens, but limits joint tokens to only attend with tokens belonging to the same human. In the full attention setting, we allow tokens to any other tokens. The inter-person setting will further limit the hand and face location tokens to attend with only its subject. As shown in \Tab~\ref{tab:abl2}, the unlimited setting is the worst, as the complicated attention mechanism is not properly learned. And the limited setting is also not ideal compared to our original attention mechanism. Furthermore, we visualize the cross-attention of our model. As shown in \Fig\ref{fig:attention}, our model is able to localize global features with body location tokens and local features with joint tokens. The lower part shows the attention map under occlusion, and it shows that our model will take reference from other body parts.

%% file: tables/agora_test.tex
\begin{table*}[h]

  \centering
  \vspace{-2mm}
  \resizebox{\textwidth}{!}{
  \begin{tabular}{lccccccccccccccccc}
    \toprule

    \multirow{2}{*}{Methods} & 
    \multirow{2}{*}{F Score$\uparrow$}& \multirow{2}{*}{Precision$\uparrow$} & \multirow{2}{*}{Recall$\uparrow$} &
    \multicolumn{2}{c}{NMVE$\downarrow$ (\emph{mm})} &
    \multicolumn{2}{c}{NMJE$\downarrow$ (\emph{mm})} &
    \multicolumn{5}{c}{MVE$\downarrow$ (\emph{mm})} &
    \multicolumn{5}{c}{MPJPE$\downarrow$ (\emph{mm})} \\
    
     \cmidrule(lr){5-6} \cmidrule(lr){7-8} \cmidrule(lr){9-13} \cmidrule(lr){14-18}  
     & & & &
    All & Body &
    All & Body &
    All & Body & Face & LHand & RHand &
    All & Body & Face & LHand & RHhand 
    \\
    \midrule
    BEDLAM~\cite{bedlam}& 0.73 & \color{red}{0.98} & 0.59 &
    179.5 & 132.2 & 177.5 & 131.4 & 131.0 & 96.5 & {25.8} & {38.8} & \color{blue}{39.0} & 129.6 & 95.9 & {27.8} & {36.6} & \color{blue}{36.7} \\
    
    H4W~\cite{hand4whole}$^\dagger$& \color{red}{0.94} & \color{blue}{0.96} & \color{blue}{0.92} &
    144.1 & 96.0 & 141.1 & 92.7 & 135.5 & 90.2 & 41.6 & 46.3 & 48.1 & 132.6 & 87.1 & 46.1 & 44.3 & 46.2 \\
    
    BEDLAM~\cite{bedlam}$^\dagger$ & 0.73 & \color{red}{0.98} & 0.59 &
    142.2 & 102.1 & 
    141.0 & 101.8 & 
    {103.8} & 74.5 & \color{red}{23.1} & \color{red}{31.7} & 
    \color{red}{33.2} & {102.9} & 74.3 & \color{red}{24.7} & \color{red}{29.9} & \color{red}{31.3} \\
    
    PyMaF-X~\cite{pymafx2023}$^\dagger$ & 0.89 & 0.90 & 0.89&
    141.2 & 94.4 & 
    140.0 & 93.5 & 
    125.7 & 84.0 & 35.0 & 44.6 & 
    45.6 & 124.6 & 83.2 & 37.9 & 42.5 & 43.7 \\
    
    OSX~\cite{osx} $^\ast$&\color{red}{0.94} & \color{blue}{0.96} & \color{red}{0.93}&
    130.6 & 85.3 & 
    127.6 & 83.3 & 
    122.8 & 80.2 & 36.2 & 45.4 & 
    46.1 & 119.9 & 78.3 & 37.9 & 43.0 & 43.9 \\

    HybrIK-X~\cite{hybrikx} & \color{blue}{0.93} & 0.95 & \color{blue}{0.92} &
    {120.5} & {73.7} & 
    {115.7} & {72.3} & 
    112.1 & {68.5} & 37.0 & 46.7 & 
    47.0 & 107.6 & {67.2} & 38.5 & 41.2 & 41.4 \\

    \name~\cite{smplerx}& \color{blue}{0.93} & \color{blue}{0.96} & 0.90 &
    133.1 & 88.1 & 
    128.9 & 84.6 & 
    123.8 & 81.9 & 37.4 & 43.6 & 
    44.8 & 119.9 & 78.7 & 39.5 & 41.4 & 44.8\\
    
    \name~\cite{smplerx}$^\dagger$ & \color{blue}{0.93} & \color{blue}{0.96} & 0.90 &
    {107.2} & {68.3} & 
    {104.1} & {66.3} & 
    {99.7} & {63.5} & {29.9} & {39.1} &
    {39.5} & {96.8} & {61.7} & {31.4} & {36.7} & 37.2 \\

    {Native AiOS}& \color{blue}{0.93} & \color{red}{0.98}&{0.89}&
    \color{blue}{105.7} & \color{blue}{66.5} & 
    \color{blue}{103.9} & \color{blue}{65.8} & 
    \color{blue}{98.3} & \color{blue}{61.8} & {27.2} & {40.7} &
    {41.7} & \color{blue}{96.6} & \color{blue}{61.2} & {28.4} & {38.4} & 39.4 \\
    AiOS& \color{red}{0.94} & \color{red}{0.98} & 0.90 &
    \color{red}{97.8} & \color{red}{61.3} & 
    \color{red}{96.0} & \color{red}{60.7} & 
    \color{red}{91.9} & \color{red}{57.6} & \color{blue}{24.6} & \color{blue}{38.7} &
    {39.6} & \color{red}{90.2} & \color{red}{57.1} & \color{blue}{25.7} & \color{blue}{36.4} & 37.3 \\
    \midrule[1pt]
    OSX~\cite{osx}$^{\ast\diamond}$& \color{red}{0.96} & \color{red}{0.97} & \color{red}{0.95}&
    {126.4} & {81.8} & 
    {123.4} & {80.0} & 
    {121.3} & {78.5} & {36.1} & {45.9} &
    {46.3} & {118.5} & {76.8} & {37.6} & {43.5} & 44.0 \\

    \name~\cite{smplerx}$^{\dagger\diamond}$& \color{red}{0.96} & \color{red}{0.97} & \color{red}{0.95}&
    \color{red}{102.4} & \color{blue}{63.8} & 
    \color{red}{99.5} & \color{red}{62.1} & 
    \color{red}{98.3} & \color{blue}{61.2} & \color{blue}{30.3} & \color{red}{40.4} &
    \color{red}{40.7} & \color{red}{95.5} & \color{red}{59.6} & \color{blue}{31.7} & \color{red}{37.9} & \color{red}{38.2} \\

    {AiOS}& \color{red}{0.96} & \color{red}{0.97} & \color{red}{0.95}&
    \color{blue}{103.0} & \color{red}{63.5} & 
    \color{blue}{100.8} & \color{blue}{62.6} & 
    \color{blue}{98.9} & \color{red}{61.0} & \color{red}{27.7} & \color{blue}{42.5} &
    \color{blue}{43.4} & \color{blue}{96.8} & \color{blue}{60.1} & \color{red}{29.2} & \color{blue}{40.1} & \color{blue}{40.9} \\

    \bottomrule
  \end{tabular}}

  \vspace{-2mm}
  \caption{\textbf{AGORA SMPL-X test set.} $\dagger$ denotes the methods finetuned on the AGORA training set. $\ast$ denotes the methods trained on the AGORA training set only. $\diamond$ denotes the methods that use the AiOS's bounding box to crop the image.
  The {\color{red}best results} are colored with {\color{red} red}, and the {\color{blue} second-best results} are colored with {\color{blue} blue} for the upper and lower parts of the table, respectively.}
  \label{tab:agora_test}
  \vspace{-4mm}
\end{table*}

%% file: tables/agora_smpl_test.tex
\begin{table}[ht]
\centering
\resizebox{0.48\textwidth}{!}{
    \begin{tabular}{lccccccc}
    \toprule
    Methods & F1-score$\uparrow$ & Precision$\uparrow$ & Recall$\uparrow$ 
    &NMVE$\downarrow$ & NMJE$\downarrow$ & MVE$\downarrow$ & MPJPE$\downarrow$\\

    \midrule
    \multicolumn{8}{c}{Top-down Methods}\\
    \midrule

    HMR~\cite{hmr} & 0.80 & 0.93 & 0.70 &
    217.0 & 226.0 & 173.6 & 180.5 \\
    PyMAF~\cite{pymaf} & 0.84 & 0.86 & 0.82 &
    200.2 & 207.4 & 168.2 & 174.2 \\
    PARE~\cite{pare} & 0.84 & 0.96 & 0.75 &
    167.7 & 174.0 & 140.9 & 146.2 \\
    H4W~\cite{hand4whole}$^{\dagger}$ & \underline{0.94} & 0.96 & \underline{0.93} &
    90.2 & 95.5 & 84.8 &89.8 \\
    CLIFF~\cite{cliff}$^{\dagger}$ & 0.91 & 0.96 & 0.87 &
    83.5& 89.0 & 76.0 & 81.0 \\
    HybrIK~\cite{li2020hybrik}$^{\dagger}$ & 0.91 & 0.92 & 0.90 &
    81.2 & 84.6 & 73.9 & 77.0 \\
    ProPose~\cite{propose}$^{\dagger}$ & 0.90 & 0.91 & 0.89 &
    78.8 & 82.7 & 70.9 & 74.4 \\
    PLIKS~\cite{pliks}$^{\dagger}$ & \underline{0.94} & 0.95 & \underline{0.93} &
    71.6& 76.1 & 67.3 & 71.5 \\
    NIKI~\cite{niki}$^{\dagger}$ & 0.91 & 0.92 & 0.90 &
    70.2& 74.0 & 63.9 & 67.3 \\

    \midrule
    \multicolumn{8}{c}{One-stage Methods}\\
    \midrule
    ROMP~\cite{romp}$^{\dagger}$ & 0.91 & 0.95 & 0.88 &
    113.6 & 118.8 & 103.4 & 108.1 \\
    BEV~\cite{bev}$^{\dagger}$ & 0.93 & 0.96 & 0.90 &
    108.3 & 113.2 & 100.7 & 105.3 \\
    $\text{AiOS}_{0.5}$ & \underline{0.94} & \textbf{0.98} & 0.90 &
    \textbf{61.2}& \textbf{68.0} & \textbf{57.5} & \textbf{63.9}\\
    $\text{AiOS}_{0.3}$ & \textbf{0.96} & \underline{0.97} & \textbf{0.95}&
    \underline{63.4}& \underline{70.1} & \underline{60.9} & 67.3\\
    
    \bottomrule
    \end{tabular}}
    
\vspace{-2mm}
\caption{ \textbf{AGORA SMPL test set}. 
$\dagger$ indicates that this method is fine-tuned on the AGORA training set. $\text{AiOS}_{0.5}$ and $\text{AiOS}_{0.3}$, representing the use of a 0.5 score threshold and a 0.3 score threshold to filter the data, respectively.
}
\vspace{-4mm}
\label{tab:agora_smpl}

\end{table}

%% file: tables/ubody.tex
\begin{table}[t]
\centering
\resizebox{0.5\textwidth}{!}{
    \begin{tabular}{lcccccc}
    \toprule
    & 
    \multicolumn{3}{c}{PA-PVE$\downarrow$ (\emph{mm})} &
    \multicolumn{3}{c}{PVE$\downarrow$ (\emph{mm})} \\
    
    \cmidrule(lr){2-4} \cmidrule(lr){5-7} 
    
    Method &
    All & Hands & Face &
    All & Hands & Face \\
    
    \midrule
    PIXIE~\cite{pixie} & 
    61.7 & 12.2 & 4.2 & 168.4 & 55.6 & 45.2 \\
    
    H4W~\cite{hand4whole} & 
    44.8 & {8.9} & 2.8 & 104.1 & 45.7 & 27.0 \\

    OSX~\cite{osx} & 
    42.4 & 10.8 & \underline{2.4} & 92.4 & 47.7 & 24.9 \\

    OSX~\cite{osx}$^\dagger$ & 
    42.2 & \underline{8.6} & \textbf{2.0} & 81.9 & 41.5 & {21.2} \\
    
    
    \name~\cite{smplerx} & 
    33.2 & 10.6 & 2.8 & {61.5} & 43.3 & 23.1 \\
    
    
    \name~\cite{smplerx}$^\dagger$ & 
    \textbf{31.9} & 10.3 & 2.8 & \textbf{57.4} & \underline{40.2} & {21.6} \\
    
    

    Native AiOS & 
    {35.6} & \underline{8.6} & {2.9} & {62.7} & {41.3} & \underline{20.8} \\
    
    AiOS & 
    \underline{32.5} & \textbf{7.3} & {2.8} & \underline{58.6} & \textbf{39.0} & \textbf{19.6} \\

    \bottomrule
    \end{tabular}}
\vspace{-2mm}
\caption{\textbf{UBody.} $\dagger$ indicates the model is finetuned with the UBody training set.}
\label{tab:ubody}
\vspace{-4mm}

\end{table}

%% file: tables/ehf.tex
\begin{table}[h]

\centering
\resizebox{0.5\textwidth}{!}{
    \begin{tabular}{lcccccc}
    \toprule
    & 
    \multicolumn{3}{c}{PA-PVE$\downarrow$ (\emph{mm})} &
    \multicolumn{3}{c}{PVE$\downarrow$ (\emph{mm})} \\
    
    \cmidrule(lr){2-4} \cmidrule(lr){5-7} 
    
    Method &
    All & Hands & Face &
    All & Hands & Face \\
    
    \midrule
    H4W~\cite{hand4whole} & 
    50.3 & \textbf{10.8} & 5.8 & 76.8 & \textbf{39.8} & 26.1\\
    
    OSX~\cite{osx} & 
    48.7 & 15.9 & 6.0 & 70.8 & 53.7 & 26.4\\

    \name~\cite{smplerx} & 
    \underline{37.8} & 15.0 & {5.1}& {65.4} & {49.4} & {17.4}\\
    
    
    Native AiOS & 
    {38.8} & {13.8} & \underline{4.0}& \underline{50.2} & {49.8} & \underline{17.3}\\  
    
    AiOS & 
    \textbf{34.0} & \underline{12.8} & \textbf{3.8}& \textbf{45.4} & \underline{44.1} & \textbf{16.9}\\    
    \bottomrule
    \end{tabular}
}
\vspace{-2mm}
\caption{\textbf{EHF}. 
As EHF is absent from our training data, it serves as a valuable tool to assess the generalization ability of our models.
\vspace{-4mm}
}
\label{tab:ehf}
\end{table}

%% file: tables/ablation_2.tex
\begin{table}[t]
\centering
\resizebox{0.5\textwidth}{!}{
    \begin{tabular}{lcccccc}
    \toprule
    \multirow{2}{*}{Ablation Studies}& 
    \multicolumn{3}{c}{PA-PVE$\downarrow$ (\emph{mm})} &
    \multicolumn{3}{c}{PVE$\downarrow$ (\emph{mm})} \\

    \cmidrule(lr){2-7}
     &
    All & Hands & Face &
    All & Hands & Face \\
\midrule
   \multicolumn{7}{c}{Attention Format} \\
    \midrule
    
    Full & 
    {42.5} & {7.2} & {4.2}& {54.8} & {39.0} & {25.8}\\ 

    Inter-human Only& 
    41.7 & 7.3 & 4.2 & 52.8 & 38.9 & 24.5\\ 

    Ours& 
   \textbf{39.9} & \textbf{7.2} & \textbf{4.1}&  \textbf{50.5} & \textbf{37.4} & \textbf{23.3}\\ 
    
    \midrule
    \multicolumn{7}{c}{ SMPL-X Supervision Manners }\\
    \midrule
    All stages & 
    42.7 & 7.4 & 4.2& 55.7 & 39.8 & 25.1\\ 
    3rd stage only & 
    40.3& 7.2 & 4.2& 51.8 & 38.0 & 23.8\\ 
    Ours (2,3 stage) & 
    \textbf{39.9} & \textbf{7.2} & \textbf{4.1}&  \textbf{50.5} & \textbf{37.4} & \textbf{23.3}\\
   \bottomrule
    \end{tabular}
    
}
\vspace{-2mm}
\caption{\textbf{Ablation Studies}. The upper part studies the attention format, and the bottom part studies the SMPL-X supervision manners.
}
\vspace{-10pt}
\label{tab:abl2}
\end{table}

%% file: sections/5_conclusions.tex
\section{Conclusion}
In this work, we propose the first all-in-one-stage model for expressive human pose and shape estimation.
We explored the incorporation of body-, face-, and hand-related tokens, as well as the aggregation of local and global features with various supervision. Moreover, we carefully designed a self-attention mechanism to establish the associations between inter- and intra-human body and body parts, which helps us to achieve its best performance. The SOTA results indicate our one-stage pipeline, which treats EHPS as a progressive set prediction problem with various sequential detections following the DETR, is a crucial factor contributing to the overall performance. This can be further proved by the performance of our naive AiOS baseline. We hope this work can contribute new insights to the EHPS research community.

\noindent\textbf{Limitations.} First, our model achieves SOTA, but there is still a large room for improvement if we add more datasets for training, particularly those containing multi-person real data. Second, the versatile design can be further extended with more dimensions of human perception tasks such as tracking and 3D localization. Exploring the estimation of hands under limited resolution is also worth investigating. 

\noindent\textbf{Acknowledgement.} This project is supported by the Hong Kong Innovation and Technology Commission (InnoHK Project CIMDA). It is also supported by the Ministry of Education, Singapore, under its MOE AcRF Tier 2 (MOET2EP20221- 0012), NTU NAP, and under the RIE2020 Industry Alignment Fund – Industry Collaboration Projects (IAF-ICP) Funding Initiative, as well as cash and in-kind contribution from the industry partner(s).

%% file: sections/A_overview.tex
\section{Overview}

Due to the limited space on the main paper, we present more additional details in this supplementary material, covering the following aspects:
\begin{itemize}
    \item Additional details on datasets, experimental parameters related to data augmentation, and implementation details in \Sec~\ref{sec_1}.
    \item Supplementary experiments investigating the impact of bounding boxes on algorithm performance in \Sec~\ref{sec_2}. 
    \item Extra SOTA quantitative and qualitative comparison experiments on EgoBody-EgoSet~\cite{egobody}, BEDLAM~\cite{bedlam}, and ARCTIC~\cite{arctic}, and extra visualization comparison on single person, multiperson, and synthetic images. in \Sec~\ref{sec_4}. 
\end{itemize}

%% file: sections/B_datasets.tex
\begin{figure*}[h]
    \centering
  \includegraphics[width=1\linewidth]{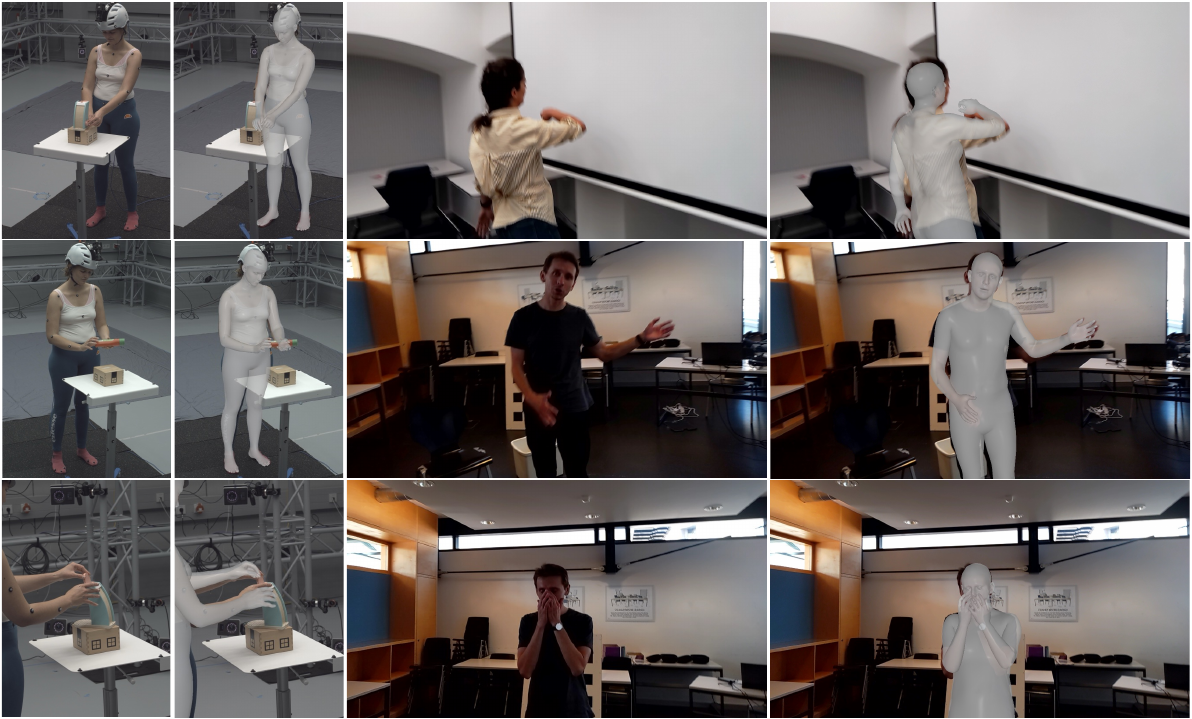}
    \caption{\textbf{Illustration of AiOS in indoor datasets.}
    The first two columns are the qualitative results on ARCTIC~\cite{arctic}, while the last two columns are the qualitative results on Egobody~\cite{egobody}.} 
    \label{fig:sup_arctic}
\end{figure*}

\begin{figure*}[h]
    \centering
  \includegraphics[width=1\linewidth]{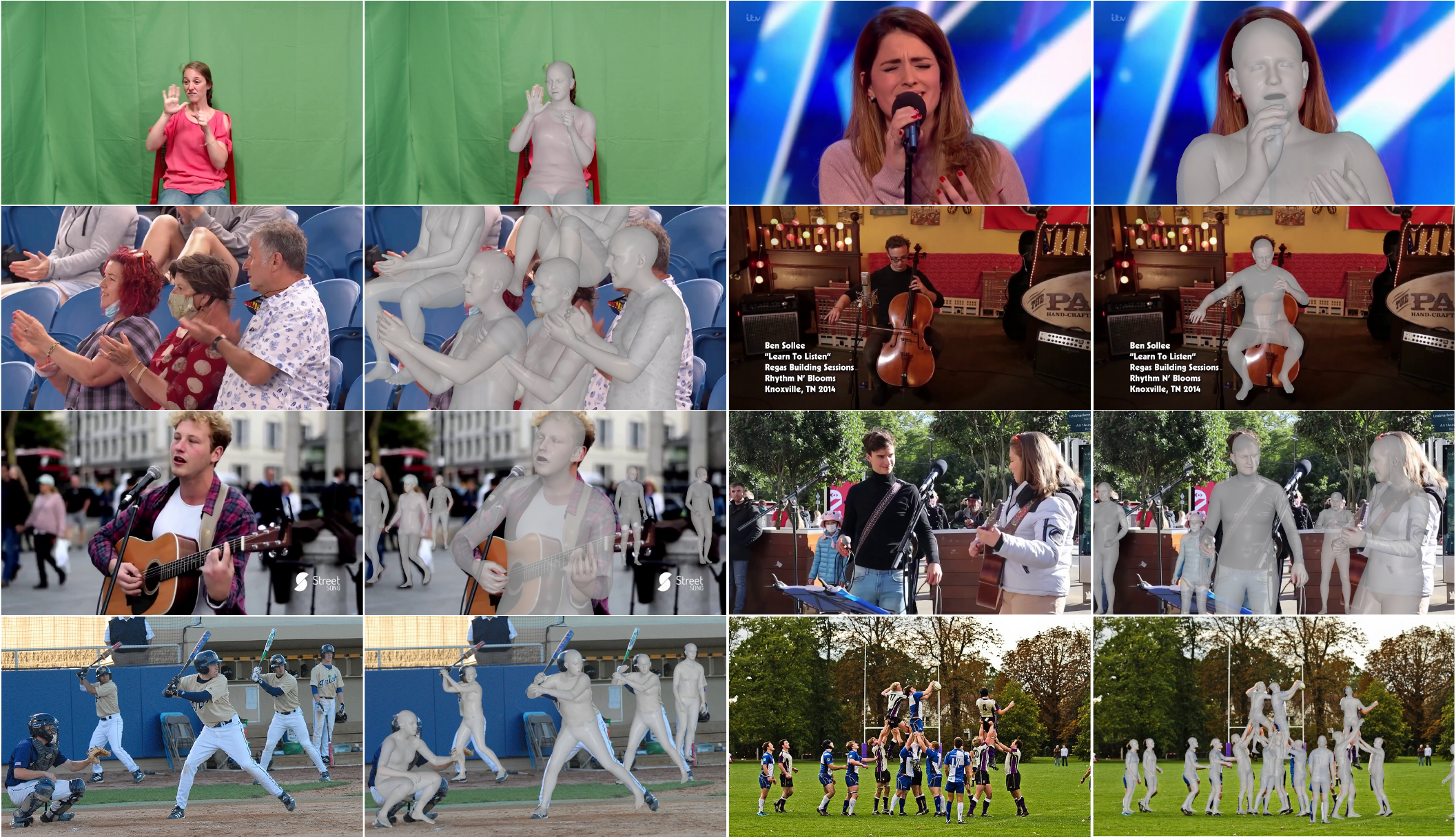}
    \caption{\textbf{Illustration of AiOS in outdoor datasets.} 
    The first three rows are qualitative results on UBody~\cite{osx}, and the last row is qualitative results on COCO~\cite{coco} }
    \label{fig:sup_ubody_new}
\end{figure*}

\section{Experiment Setup}\label{sec_1}
\subsection{Datasets}
This subsection primarily describes the characteristics of the datasets utilized in our main paper and provides details on how we use these datasets in the training and testing stages.

\noindent
\textbf{AGORA}~\cite{agora} is a synthetic image dataset that encompasses lots of complex scenarios, including severe occlusion and truncations. It has 14K training images with 122K instances and 1K validation images with around 8K instances. Recently, AGORA has become an essential benchmark for tasks related to SMPL~\cite{smpl} and SMPL-X~\cite{smplx}, primarily for its effectiveness in assessing algorithm performance in occlusion-heavy scenes. Our approach utilizes the complete dataset for training, validation, and testing purposes.

\noindent
\textbf{BEDLAM}~\cite{bedlam} is a synthetic video dataset offering a wide variety of data, including diverse body shapes, motions, skin tones, hairstyles, and clothing. The clothing is notably rendered using a professional physics simulator, enhancing realism, particularly in depicting character movement. Each image in the dataset features between 1 to 10 individuals. Originally comprising 286K images, we downscaled it by a factor of 5, yielding 57L images with 190K instances, which we then utilized for training.

\noindent
\textbf{MSCOCO}~\cite{coco} is a large-scale real-world dataset designed for object detection, segmentation, keypoint detection, and captioning. On the basis of the keypoint subset, we use the pseudo SMPL-X~\cite{smplx} annotations from NeuralAnnot~\cite{moon2022neuralannot} for training. It contains 56K multi-person images, featuring a total of 147K instances.

\noindent
\textbf{EHF}~\cite{smplx} is an indoor dataset consisting of only 100 images along with corresponding SMPL-X labels. Due to the absence of a corresponding training set, current algorithms often utilize it to assess algorithm generalization. It only contains test sets with 100 single-person images.
 
 \noindent
 \textbf{UBody}~\cite{osx} is a dataset containing diverse real-life scenarios, including movies, TV shows, talk shows, vlogs, sign language, online classes, and more. UBody contains an extensive collection of rich gestures and expressions that are not present in other real-world datasets. We down-sample the training set to 54K images with 66K instances. We utilize a downsampled test set, as used in SMPLer-X\cite{smplerx} and OSX~\cite{osx}, which has 2642 images with 2642 instances.

 \noindent
\textbf{ARCTIC}~\cite{arctic} is an indoor dataset which mainly focuses on the hand-object interaction. We downsample the original training set 1000 times for a train set of 50K images with 50K instances. We discard the egocentric view, which only contains hands, in our training. For the test set, following~\cite{smplerx}, we keep the original test set with 207K images to evaluate our method.
 
 \noindent
 \textbf{Egobody}~\cite{egobody} is a large-scale dataset for egocentric views. The egocentric view datasets are collected using Microsoft HoloLens 2, including RGB, depth, eye gaze, head tracking, and hand tracking. The SMPL-X data is obtained by fitting the multi-Kinect rig data. We downsample 2 times to get 45K images with 45K instances in the egocentric-view split for training. The test set has 62K images.

We pre-train our model by randomly sampling data from AGORA, BEDLAM, and COCO. We choose these three datasets because their multi-person images satisfy our method of perceiving the positions of individuals in the images.
Additionally, AGORA and BEDLAM have the advantage of accurate ground truth, while COCO provides the diversity of real-world scenes, preventing our method from overfitting to synthetic data.

SMPLer-X~\cite{smplerx} indicates that scaling up the data can enhance algorithm performance. Hence, we incorporate Egobody, ARCTIC, and UBody data into our training. The sampling probabilities are 0.2, 0.2, 0.2, 0.2, 0.1, and 0.1, respectively, for AGORA, BEDLAM, COCO, Egobody, ARCTIC, and UBody. ARCTIC and Egobody provide more accurate real-world whole-body data compared to COCO, while UBody contributes abundant and diverse gestures and expressions. All the datasets are standardized into the HumanData~\cite{mmhuman3d} format. Some visual demonstrations of AiOS are provided in \Fig\ref{fig:sup_arctic} and \Fig\ref{fig:sup_ubody_new}.

\subsection{Implementation details}
The training is conducted on 16 V100 GPUs, with a total batch size of 32. 
We initialize AiOS's backbone, encoder, and part of the decoder from a weight trained on COCO human pose estimation, provided by ED-Pose ~\cite{edpose}. We use Adam optimizer with a step learning rate for training. We first train our model for 60 epochs with an initial learning rate $1 \times 10^{-4}$ and drop at the 50th epoch on AGORA, BEDLAM, and COCO. We finetune it for 50 epochs with $1 \times 10^{-5}$ initial learning rate and drop at the 25th epoch on all train datasets.

During the training stage, we adopt color jittering, random horizontal flipping, random image resizing, and random instance cropping.
For the color jittering, we randomly apply a variation of $\pm0.2$ in the RGB channels. For the random horizontal flipping, we augment images and their corresponding annotations by horizontally flipping them with a probability of 0.5. For random image resizing, during the training process, we resize the images in proportion, ensuring that the shorter side is kept between 480 and 800 pixels and the longer side does not exceed 1333 pixels. During testing, we set the shorter side to 800 pixels, and the longer side is scaled proportionally, with the constraint that it does not exceed 1333 pixels. For random instance cropping, we first randomly enable the cropping operation with a probability of 0.5. When performing the cropping operation on a multi-person dataset, such as AGORA~\cite{agora}, BEDLAM~\cite{bedlam}, COCO~\cite{coco}, We randomly sample $1$ to $N$ instances from the image with probability 0.5, where N is the total number of people in the image. The image is then cropped according to their collective bounding box. Alternatively, with a 0.5 probability, the cropping operation is applied to the collective bounding box of all instances. We maintain the original aspect ratio during the cropping process to avoid cropping unusual aspect ratios.

\subsection{Loss Functions}

\noindent\textbf{Body-location Decoder}. The losses for supervising Body-location Decoders are $L_{box}$ and $L_{cls}$. $L_{box}$ contains L1 loss and the GIOU loss~\cite{giou} for the body location. $L_{cls}$ is 
focal loss ~\cite{lin2017focal} for classify body tokens.

\noindent\textbf{Body-refinement Decoder.}. The losses for supervising Body-refinement Decoders are $L_{box}$, $L_{j2d}$ and $L_{smplx_b}$. $L_{box}$ contains L1 and loss GIOU loss for the body location, hands location, and face location. $L_{j2d}$ is the L1 loss and OKS loss supervising the body joints location and $L_{smplx}$ contains L1 loss with ground truth SMPL-X body parameters $L_{param}$, L1 loss $L_{kp3d}$ for 3D body keypoints regressed by SMPL-X J-regressor, and L1 loss $L_{kp2d}$ for projected 2D keypoints. 

\noindent\textbf{Wholebody-refinement Decoder.}. The losses for supervising Wholebody-refinement Decoders are $L_{j2d}$ and $L_{smplx}$.  $L_{j2d}$ is the L1 loss and OKS loss supervising the whole-body joints location and $L_{smplx}$ contains L1 loss with ground truth SMPL-X parameters $L_{param}$, L1 loss $L_{kp3d}$ for 3D whole-body keypoints regressed by SMPL-X J-regressor, and L1 loss $L_{kp2d}$ for projected 2D whole-body keypoints. 

\noindent\textbf{Loss Weights} We weighted-sum all the losses at all stages as the final loss. The loss weights of the same type of loss in different stages are the same, which are shown as follows: $L_{cls}$: 2.0, $L_{smplx}$: 1.0 (pose), 0.01 (shape, expression), $L_{kp3d}$: 1.0 (body), 0.5 (face, hand), $L_{kp2d}$: 1.0 (body), 0.5 (face, hand), $L^1_{j2d}$: 10, $L_{oks}$: 4.0 (body), 0.5 (face), 0.5 (hands),  $L_{giou}$: 2.0, $L^1_{box}$: 5.0.

%% file: sections/C_experiment_1.tex
\section{Sensitivity Analysis of Performance to Bounding Box Accuracy}\label{sec_2}
In this section, we present that current methodologies exhibit a significant sensitivity to bounding boxes. Our experiments are carried out on the AGORA~\cite{agora} validation set, utilizing official evaluation tools~\footnote{{https://github.com/pixelite1201/agora\_evaluation}} for a thorough assessment. We report several key metrics, including Normalized Mean Vertex Error (NMVE), Normalized Mean Joint Error (NMJE), Mean Vertex Error (MVE), and Mean Per-Joint Position Error (MPJPE), which focus on the reconstruction accuracy of body, hands, and face. Additionally, we include F-Score, precision, and recall to evaluate the accuracy of detection.

\begin{figure}[th]
\vspace{-0.3cm}
  \centering
  \includegraphics[width=1\linewidth]{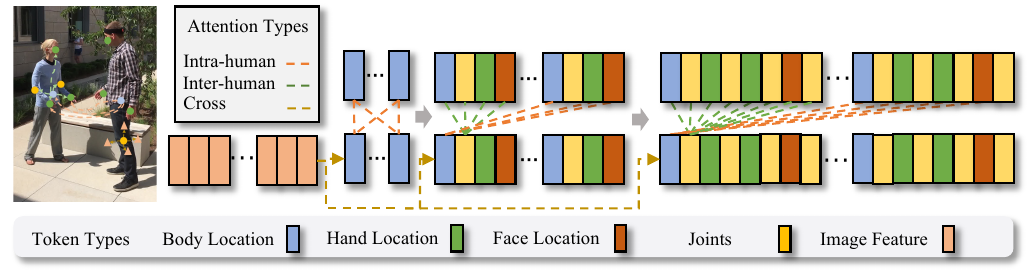}
  \vspace{-2mm}
   \caption{\textbf{Illustration of Inter-human, Intra-human, and Cross Attention.} Inter-human attention is conducted between the body tokens and body, hand, and face location tokens of different persons'. Intra-human attention is conducted with the location and joint tokens of the same person. Cross-attention is conducted between image features and all the tokens.}
   \vspace{-5mm}
   \label{fig:atten}
\end{figure}

\input{tables/agora_val}

Firstly, we utilize the ground truth (GT) bounding box to crop the image and evaluate the performance of OSX~\cite{osx} and SMPLer-X~\cite{smplerx}, denoted by \textbf{GT Box} in  \Tab~\ref{tab:agora_val}.
Following this, we employ DAB-DETR\cite{liu2022dab, mmdet}, an off-the-shelf detection model, to identify human bounding boxes, replacing the GT boxes for image cropping.
We present detection results under three bounding box score thresholds: 0.1, 0.2, and 0.3, labeled as detected boxes with scores 0.1, 0.2, and 0.3, respectively. A higher score indicates greater confidence in the detection results, but it may lead to missing instances with severe occlusion. This is reflected in the metrics as high accuracy but low recall. 
Conversely, a lower score threshold retains results with lower confidence, capturing more instances with severe occlusion but resulting in many redundant detections. This is reflected in the metrics as high recall but lower accuracy. 


From the data presented in \Tab~\ref{tab:agora_val}, it's evident that regardless of using a low or high threshold for filtering detection results, both OSX~\cite{osx} and SMPLer-X~\cite{smplerx} show a marked decrease in performance in terms of NVME and NMJE when compared to results achieved using GT bounding boxes. This performance gap is largely due to the way NMVE and NMJE are normalized using the F1 score. The F1 score, being the harmonic mean of recall and precision, penalizes methods for both missed detections and false positives, which explains the observed discrepancy in performance.

Notably, when we filter the detection results with a score of 0.1, denoted by \textbf{Detected box \textit{w} score 0.1}, the recall is 0.92, indicating that we detect the majority of instances, although there were many redundant detections. In this case, if we concentrate on reconstruction accuracy, represented by MVE and MPJPE, we observe that the results using detected bounding boxes OSX and SMPLer-X are significantly worse than using GT bounding box for cropping images.

On the lower part of \Tab~\ref{tab:agora_val}, we present the AiOS results. To allow more hard cases to be detected, we use a threshold of 0.1 to filter the result. Similar to the main paper, providing AiOS's bounding boxes to OSX and SMPLer-X denoted by \textbf{AiOS box}, leads to a significant performance increase compared to using the detected boxes (\textbf{Detected box \textit{w} score 0.3}), even though AiOS's box includes more challenging cases. 
Specifically, there is a significant improvement in NMVE with a 16\% reduction from 148.1 to 124.3 for OSX and an 18\% decrease from 126.4 to 103.3 for SMPLer-X. 
Additionally, there is an improvement in NMJE, with a 15\% reduction from 142.1 to 120.0 for OSX and a 17\% decrease from 121.3 to 99.6 for SMPLer-X.
These results substantiate that one-stage methods demonstrate superior performance in real-world scenarios compared to existing two-stage methods.

\begin{figure*}[h]
    \centering
  \includegraphics[width=1\linewidth]{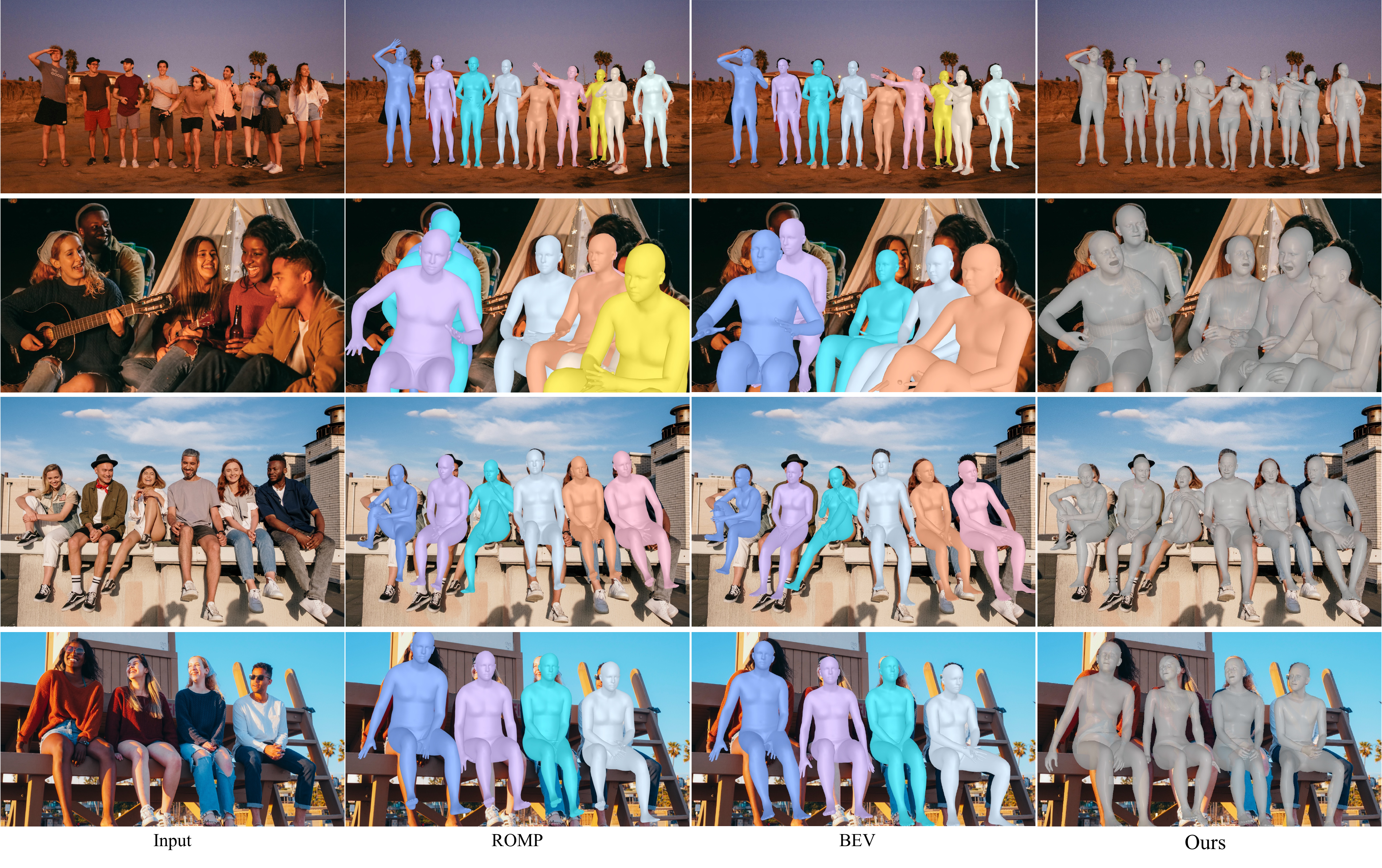}
  \includegraphics[width=1\linewidth]{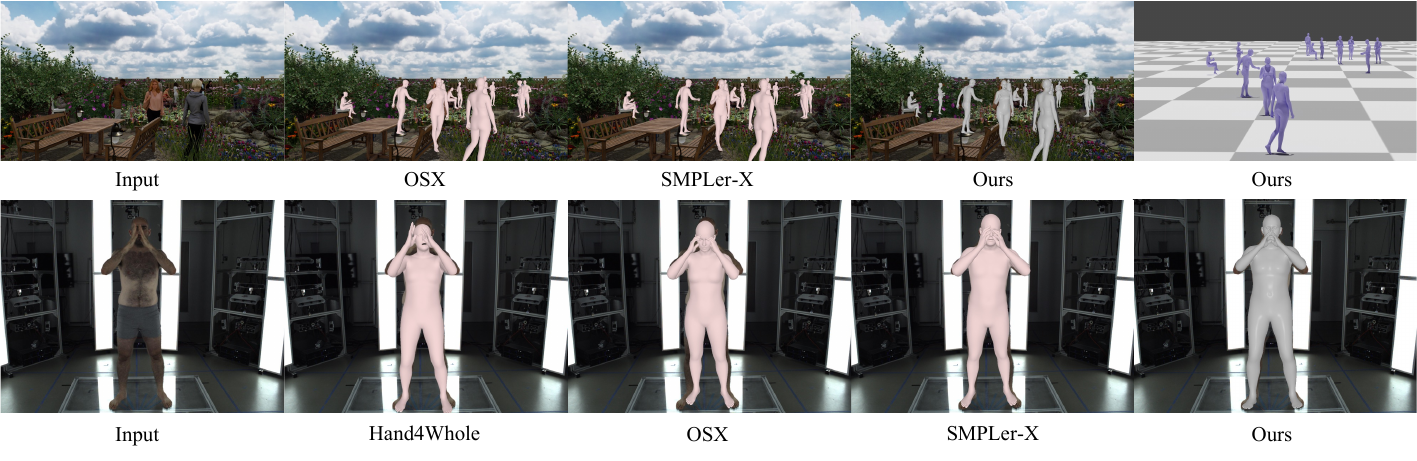}
    \caption{\textbf{Additional visual comparisons with existing methods.} 
    \textbf{Upper part:} The first column is the input images, and they are downloaded from the internet. The second column is the visualization results of ROMP~\cite{romp}, the third column shows the visualization results of BEV~\cite{bev}, and the last column illustrates our visualization results; \textbf{Middle part:} Comparison of current SOTA methods ~\cite{smplerx,osx} with our AiOS model on AGORA~\cite{agora}. \textbf{Lower part:}Comparison of current SOTA methods ~\cite{smplerx,hand4whole,osx} with our AiOS model on EHF test~\cite{expose}.
    }
    \label{fig:sup_romp}
\end{figure*}
\input{tables/bedlam}
\input{tables/arctic}
\input{tables/egobody}

For a fair comparison with OSX and SMPLer-X, which use the GT bounding boxes to crop images, we filter out the same instances that AiOS failed to detect for both OSX and SMPLer-X. Interestingly, we found that the results obtained using AiOS bounding boxes (denoted with \textbf{AiOS box} in \Tab~\ref{tab:agora_val}) are comparable to those obtained using GT bounding boxes (denoted with \textbf{GT box{$^\times$}}). Notably, under this bounding box setting, AiOS still outperforms the current state-of-the-art OSX, even though it utilizes the GT bounding box and is on par with the foundational model SMPler-X L20.
To further demonstrate the superiority of AiOS, we follow the procedure in RoboSMPLX~\cite{robosmplx} to translate the image by 0.1 of the image size horizontally, introducing a small noise to the GT bounding boxes (denoted by \textbf{GT box with noise$^\times$} in \Tab~\ref{tab:agora_val_noise}). It's worth noting that this noise only shifts the person away from the image center and does not remove the person from the image plane, avoiding truncation and occlusion. Firstly, compare it with the results cropping with the GT bounding box denoted by \textbf{GT box$^\times$}, we can observe a drop in performance. Specifically, for NMVE, OSX increased from 123.6  to 126.1, and SMPLer-X increased from 101.5  to 105.6. Regarding NMJE, OSX increased from 119.3  to 121.8, and SMPLer-X increased from 97.6  to 101.6. This observation indicates that current two-stage methods are highly sensitive to the accuracy of the bounding box, as even a slight noise introduced, causing the person to be off-center, results in a performance drop, even with GT bounding boxes. 

When comparing the results obtained by cropping images with bounding boxes provided by AiOS to those obtained by cropping with GT bounding boxes and GT bounding boxes with added noise, we observe that the results of cropping with AiOS-provided bounding boxes are slightly inferior to those obtained with GT bounding boxes in body-related metrics but better than GT bounding boxes with added noise. However, for the face and hands, the results of cropping with AiOS-provided bounding boxes can even be better than those obtained with GT bounding boxes. We attribute this improvement to AiOS's attention to not only the body but also the hands and face.

%% file: tables/agora_val.tex
\begin{table*}[t]

  \centering
  \vspace{-2mm}
  \resizebox{\textwidth}{!}{
  \begin{tabular}{llccccccccccccccccc}
    \toprule

    \multirow{2}{*}{Methods} &\multirow{2}{*}{Box Type} &
    \multirow{2}{*}{F Score$\uparrow$}& \multirow{2}{*}{Precision$\uparrow$} & \multirow{2}{*}{Recall$\uparrow$} &
    \multicolumn{2}{c}{NMVE$\downarrow$ (\emph{mm})} &
    \multicolumn{2}{c}{NMJE$\downarrow$ (\emph{mm})} &
    \multicolumn{5}{c}{MVE$\downarrow$ (\emph{mm})} &
    \multicolumn{5}{c}{MPJPE$\downarrow$ (\emph{mm})} \\
    
     \cmidrule(lr){6-7} \cmidrule(lr){8-9} \cmidrule(lr){10-14} \cmidrule(lr){15-19}  
     & & & & &
    All & Body &
    All & Body &
    All & Body & Face & LHand & RHand &
    All & Body & Face & LHand & RHhand 
    \\

    \midrule
     \multirow{8}{*}{OSX~\cite{osx}}&GT box& {\color{red}1.0}&{\color{red}1.0}&{\color{red}1.0}&
     {\color{red}120.4}&{\color{red}75.2}&{\color{red}116.1}&{\color{red}72.1}&
     {120.4}&{75.2}&{39.4}&{47.5}&{48.8}&
     {116.1}&{72.1}&{40.5}&{45.1}&{46.4} \\
     

     
     &Detected box \textit{w} score 0.1&0.58&0.42&0.92&
     237.8&155.9&228.3&147.4&137.9&90.4&43.8&48.3&50.4&132.4&85.5&46.4&46.1&48.1 \\
     &Detected box \textit{w} score 0.2 &0.8&0.74&0.87&
     165.9&110.6&159.2&104.7&
     132.7&88.5&40.7&{\color{blue}44.9}&{\color{blue}46.9}&
     127.4&83.8&43.1&42.9&44.8 \\
     
     &Detected box \textit{w} score 0.3 &0.86&0.89&0.83&
     148.1&100.3&142.1&95.1&
     127.4&86.3&{\color{blue} 37.8}& {\color{red}41.8} &{\color{red}43.6}&
     122.2&81.8&39.8&{\color{red}39.8}&{\color{red}41.6} \\ 
     
     &GT box{$^\times$} &{\color{blue}0.95}&{\color{blue}0.93}&{\color{blue}0.97}&
     {\color{blue}123.6}& {\color{blue}77.3}&{\color{blue}119.3}& {\color{blue}74.2} & 
     {\color{red}117.4} &{\color{red}73.4}&38.2&46.3&47.7&
     {\color{red}113.3}&{\color{red}70.5}&{\color{blue}39.0}&43.9&45.3\\
     
     &AiOS box&{\color{blue}0.95}&{\color{blue}0.93}&{\color{blue}0.97}&
     124.3&78.7&120.0&75.7&
     {\color{blue}118.1}&{\color{blue}74.8}&{\color{red} 37.4}&45.5&47.0&
     {\color{blue}114.0}&{\color{blue}71.9}&{\color{red}38.3}&{\color{blue}43.2}&{\color{blue}44.7} \\
    \midrule

    \multirow{7}{*}{SMPLer-X~\cite{smplerx}}&GT box&{\color{red}1.0}&{\color{red}1.0}&{\color{red}1.0}&
    {\color{red}99.5}&{\color{red}60.2}&{\color{red}95.5}&{\color{red}57.5}&
    99.5&{\color{blue}60.2}&32.9&41.9&43.0&
    95.5&{\color{blue}57.5}&34.1&39.5&40.5 \\
    
    &Detected box \textit{w} score 0.1 &0.58&0.43&0.92&
    203.4&131.6&195.3&124.7&
    118.0&76.3&37.1&43.6&44.4&
    113.3&72.3&39.5&41.3&42.2 \\
    
    &Detected box \textit{w} score 0.2&0.81&0.75&0.88&
    140.7&92.6&135.1&87.8&
    114.0&75.0&34.7&40.7&{\color{blue}41.7}&
    109.4&71.1&36.9&38.5&39.5 \\
    
    &Detected box \textit{w} score 0.3&0.86&0.9&0.83&
    126.4&84.4&121.3&80.1&
    108.7&72.6&31.8&{\color{red}37.8}&{\color{red}38.7}&
    104.3&68.9&33.7&{\color{red}35.7}&{\color{red}36.7 }\\
    
    &GT box{$^\times$}&{\color{blue}0.95}&{\color{blue}0.93}&{\color{blue}0.97}&
    {\color{blue}101.5}& {\color{blue}61.4}& {\color{blue}97.6}& {\color{blue}58.8}&
    {\color{red}96.4}&{\color{red}58.3}&{\color{blue}31.7}&40.8&41.9&
    {\color{red}92.7}&{\color{red}55.9}&{\color{blue}32.6}&{38.3}&{\color{blue}39.4}\\
    
    &AiOS box&{\color{blue}0.95}&{\color{blue}0.93}&{\color{blue}0.97}&
    103.3&63.5&99.6&61.1&
    {\color{blue}98.1}&60.3&{\color{red}31.3}&{\color{blue}40.4}&41.8&
    {\color{blue}94.6}&58.0&{\color{red}32.3}&{\color{blue}38.0}&{\color{blue}39.4} \\

\midrule

     AiOS& \qquad\qquad\quad - &{0.95}&{0.93}&{0.97}&
     106.4&64.2&103.4&62.1&101.1&61.0&30.7&43.9&45.7&98.2&59.0&32.8&41.5&43.4 \\

    \bottomrule
  \end{tabular}}

  \vspace{-2mm}
  \caption{\textbf{AGORA validation set.} OSX~\cite{osx} and SMPLer-X~\cite{smplerx} are finetuned on the AGORA training set. However, AiOS is not intentionally fine-tuned exclusively on AGORA. \textbf{GT Box} means that this method uses the ground truth bounding box to crop the image. \textbf{GT box{$^\times$}} means that this method uses the ground truth bounding box to crop the image but filters the instances that AiOS fails to detect. \textbf{AiOS box} means that this method uses the bounding box provided by AiOS. The {\color{red}best results} are colored with {\color{red} red}, and the {\color{blue} second-best results} are colored with {\color{blue} blue} for OSX and SMPLer-X, respectively.}
  \label{tab:agora_val}
\end{table*}

\begin{table*}[t]

  \centering
  \vspace{-2mm}
  \resizebox{\textwidth}{!}{
  \begin{tabular}{llccccccccccccccccc}
    \toprule

    \multirow{2}{*}{Methods} &\multirow{2}{*}{Box Type} &
    \multirow{2}{*}{F Score$\uparrow$}& \multirow{2}{*}{Precision$\uparrow$} & \multirow{2}{*}{Recall$\uparrow$} &
    \multicolumn{2}{c}{NMVE$\downarrow$ (\emph{mm})} &
    \multicolumn{2}{c}{NMJE$\downarrow$ (\emph{mm})} &
    \multicolumn{5}{c}{MVE$\downarrow$ (\emph{mm})} &
    \multicolumn{5}{c}{MPJPE$\downarrow$ (\emph{mm})} \\
    
     \cmidrule(lr){6-7} \cmidrule(lr){8-9} \cmidrule(lr){10-14} \cmidrule(lr){15-19}  
     & & & & &
    All & Body &
    All & Body &
    All & Body & Face & LHand & RHand &
    All & Body & Face & LHand & RHhand 
    \\

    \midrule
     \multirow{4}{*}{OSX~\cite{osx}}

      &GT box{$^\times$} &{0.95}&{0.93}&{0.97}&
     {\color{red}123.6}& {\color{red}77.3}&{\color{red}119.3}& {\color{red}74.2} & 
     {\color{red}117.4} &{\color{red}73.4}&{\color{blue}38.2}&{\color{blue}46.3}&{\color{blue}47.7}&
     {\color{red}113.3}&{\color{red}70.5}&{\color{blue}39.0}&{\color{blue}43.9}&{\color{blue}45.3}\\

     &GT box with noise{$^\times$} &{0.95}&{0.93}&{0.97}&
     126.1& 78.8 & 121.8 & {\color{blue}75.6} & 
     {119.8} &{74.9}&{38.7}&{46.9}&49.1&
     {115.7}&{72.1}&{39.6}&{44.5}&46.6\\
     
     &AiOS box&{0.95}&{0.93}&{0.97}&
     {\color{blue}124.3}&{\color{blue}78.7}&{\color{blue}120.0}&{75.7}&
     {\color{blue}118.1}&{\color{blue}74.8}&{\color{red} 37.4}&{\color{red}45.5}&{\color{red}47.0}&
     {\color{blue}114.0}&{\color{blue}71.9}&{\color{red}38.3}&{\color{red}43.2}&{\color{red}44.7} \\
    \midrule

    \multirow{4}{*}{SMPLer-X~\cite{smplerx}}

    &GT box{$^\times$}&{0.95}&{0.93}&{0.97}&
    {\color{red}101.5}& {\color{red}61.4}& {\color{red}97.6}& {\color{red}58.8}&
    {\color{red}96.4}&{\color{red}58.3}&{\color{blue}31.7}&{\color{blue}40.8}&{\color{blue}41.9}&
    {\color{red}92.7}&{\color{red}55.9}&{\color{blue}32.6}&{\color{blue}38.3}&{\color{red}39.4}\\

    &GT box with noise{$^\times$} &{0.95}&{0.93}&{0.97}&
    105.6& 64.0 & 101.6& 61.5&
    100.3&60.8&{32.5}&42.4&43.6&
    96.5&58.4&{33.5}&39.9&{\color{blue}41.0}\\

    &AiOS box&{0.95}&{0.93}&{0.97}&
    {\color{blue}103.3}&63.5&{\color{blue}99.6}&{\color{blue}61.1}&
    {\color{blue}98.1}&{\color{blue}60.3}&{\color{red}31.3}&{\color{red}40.4}&{\color{red}41.8}&
    {\color{blue}94.6}&{\color{blue}58.0}&{\color{red}32.3}&{\color{red}38.0}&{\color{red}39.4} \\

    \bottomrule
  \end{tabular}}

  \vspace{-2mm}
  \caption{\textbf{AGORA validation set with noise bounding box.} OSX~\cite{osx} and SMPLer-X~\cite{smplerx} are finetuned on the AGORA training set. 
  \textbf{GT box} means that this method uses the ground truth bounding box to crop the image.
  \textbf{GT box with noise} means translating the ground truth bounding box by \textbf{10\%} of the image size in the horizontal direction, causing the human to deviate from the image center. This is a very small noise that ensures the person is not removed from the image plane, avoiding truncation.
  \textbf{{$\times$}} means that this method filters the instances that AiOS fails to detect.
  The {\color{red}best results} are colored with {\color{red} red}, and the {\color{blue} second-best results} are colored with {\color{blue} blue} for OSX and SMPLer-X, respectively.}
  \label{tab:agora_val_noise}
  \vspace{-5mm}
\end{table*}

%% file: tables/bedlam.tex
\begin{table*}[ht]

  \vspace{-2mm}
  \resizebox{\textwidth}{!}{
  \begin{tabular}{lccccccccccccccccc}
    \toprule

    \multirow{2}{*}{Methods} & 
    \multirow{2}{*}{F Score$\uparrow$}& \multirow{2}{*}{Precision$\uparrow$} & \multirow{2}{*}{Recall$\uparrow$} &
    \multicolumn{2}{c}{NMVE$\downarrow$ (\emph{mm})} &
    \multicolumn{2}{c}{NMJE$\downarrow$ (\emph{mm})} &
    \multicolumn{5}{c}{MVE$\downarrow$ (\emph{mm})} &
    \multicolumn{5}{c}{MPJPE$\downarrow$ (\emph{mm})} \\
    
     \cmidrule(lr){5-6} \cmidrule(lr){7-8} \cmidrule(lr){9-13} \cmidrule(lr){14-18}  
     & & & &
    All & Body &
    All & Body &
    All & Body & Face & LHand & RHand &
    All & Body & Face & LHand & RHhand 
    \\
    \midrule

    PIXIE~\cite{pixie}& 0.94 & 0.99 & \textbf{0.90} &
     158.7 & 107.2 & 
     153.7 & 103.5 & 
     149.2 & 100.8 & 51.4 & 44.8 & 48.9
     & 144.5 & 97.3 & 55.4 & 41.3 & 44.8 \\
     
    BEDLAM-CLIFF & 0.94 & 0.99 & \textbf{0.90} &
     100.6 &  65.2 & 
     98.0 & 64.3 & 
     94.6 & 61.3 & 29.8 & 34.7 & 35.5
      & 92.1 & 60.4 & 30.4 & 32.2 & 32.6 \\

    BEDLAM-CLIFF++$^\dagger$ & 0.94 & 0.99 & \textbf{0.90} &
     93.2 & 61.2 & 
     90.9 & 60.4 & 
     87.6 & 57.5 & 27.3 & 30.3 & 32.6
     & 85.4 & 56.8 & 28.0 & 28.0 & 29.9 \\

    AiOS$^\dagger$ & \textbf{0.95} & \textbf{1} & \textbf{0.90} &
     \textbf{87.6} & \textbf{57.7} & 
     \textbf{85.8} & \textbf{57.7} & 
     \textbf{83.2} & \textbf{54.8} & \textbf{26.4} & \textbf{28.1} & \textbf{30.8}
     & \textbf{81.5} & \textbf{54.8} & \textbf{26.2} & \textbf{25.9} & \textbf{28.1}\\

    \bottomrule
  \end{tabular}}

  \vspace{-2mm}
  \caption{\textbf{BEDLAM test set.} The best results are in \textbf{bold}. $\dagger$ denotes methods
that include the AGROA training set.}
    \vspace{-5mm}
  \label{tab:bedlam_test}
\end{table*}

%% file: tables/arctic.tex
\begin{table}[h]
\centering

\resizebox{0.5\textwidth}{!}{
    \begin{tabular}{lcccccc}
    \toprule
    & 
    \multicolumn{3}{c}{PA-PVE$\downarrow$ (\emph{mm})} &
    \multicolumn{3}{c}{PVE$\downarrow$ (\emph{mm})} \\
    
    \cmidrule(lr){2-4} \cmidrule(lr){5-7} 
    
    Method &
    All & Hands & Face &
    All & Hands & Face \\
    
    \midrule
    H4W~\cite{hand4whole} & 
    63.4 & \underline{18.1} & 4.0 & 136.8 & 54.8 & 59.2 \\

    OSX~\cite{osx} & 
    56.9 & \textbf{17.5} & 3.9 & 102.6 & 56.5 & 44.6 \\
    
    OSX~\cite{osx}$^\dagger$ & 
    33.0 & 18.8 & 3.3 & 58.4 & 39.4 & 30.4 \\


    
    
    SMPLer-X~\cite{smplerx}  & {31.9} & 18.9  & {2.5} & {52.2} & \underline{39.3}& {27.0}  \\

    Native AiOS & 
    \underline{31.8} & 19.7 & \underline{2.3} & \underline{51.6} & {40.6} & \underline{26.5} \\
    AiOS & 
    \textbf{30.2} & 19.2 & \textbf{2.1} & \textbf{47.1} & \textbf{38.3} & \textbf{26.1} \\
    
    \bottomrule
    \end{tabular}}
    
\vspace{-2mm}
\caption{\textbf{ARCTIC.} $\dagger$ denotes the method finetuned on the ARCTIC training set.}
\vspace{-4mm}
\label{tab:arctic}
\end{table}

%% file: tables/egobody.tex
\begin{table}[h]
 
\centering

\resizebox{0.5\textwidth}{!}{
    \begin{tabular}{lcccccc}
    \toprule
    \multirow{2}{*}{Method}& 
    \multicolumn{3}{c}{PA-PVE$\downarrow$ (\emph{mm})} &
    \multicolumn{3}{c}{PVE$\downarrow$ (\emph{mm})} \\
    
    \cmidrule(lr){2-4} \cmidrule(lr){5-7} 
    
     &
    All & Hands & Face &
    All & Hands & Face \\
    
    \midrule
    H4W~\cite{hand4whole} & 
    58.8 & {9.7} & 3.7 & 121.9 & 50.0 & 42.5 \\

    OSX~\cite{osx} & 
    54.6 & 11.6 & 3.7 & 115.7 & 50.6 & 41.1 \\

    OSX~\cite{osx}$^\dagger$ & 
    45.3 & 10.0 & \underline{3.0} & 82.3 & 46.8 & 35.2 \\
    
    
    \name~\cite{smplerx}& 
    38.9 & 9.9 & \underline{3.0} & 66.6 & 42.7 & 31.8\\
    
    
    \name~\cite{smplerx}$^\dagger$& 
    \textbf{37.8} & 9.9 & \textbf{2.9} & \underline{63.6} & {46.3} & \underline{32.3} \\
    
    Native AiOS & 
    {40.8} & \underline{9.1} & \underline{3.0} & {64.6} & \underline{42.3} & \textbf{26.3} \\
    AiOS & 
    \underline{38.0} & \textbf{9.0} & \textbf{2.9} & \textbf{61.6} & \textbf{40.0} & \underline{26.7} \\
    \bottomrule
    \end{tabular}}

\vspace{-2mm}
\caption{\textbf{EgoBody-EgoSet.} $\dagger$ denotes the methods that are finetuned on the EgoBody-EgoSet training set. }
\vspace{-4mm}
\label{tab:egobody}

\end{table}

%% file: sections/F_experiments_3.tex
\section{Extra SOTA comparison experiments} \label{sec_4}
In this section, we show the extra single datasets on EgoBody-EgoSet, ARCTIC, and BEDLAM in \Tab~\ref{tab:egobody}, ~\ref{tab:arctic}, ~\ref{tab:bedlam_test}. Our proposed AiOS achieved SOTA performance across all these datasets. Also, we provide extra visualization comparison in~\Fig\ref{fig:sup_romp}.
